\documentclass[acmsmall, authordraft=false]{acmart}
\usepackage{changepage}
\usepackage{subcaption}
\usepackage{pdflscape}
\usepackage{graphicx}
\usepackage{multirow}
\usepackage{array}
\usepackage{tabularx}
\newcolumntype{Y}{>{\raggedright\arraybackslash}X}
\newcommand{\varindent}{\hspace{0.8em}}
\AtBeginDocument{%
  \setlength{\parindent}{3em}%
  \setlength{\normalparindent}{3em}%
}

\acmConference[XY'99]{XX Conference}{April
11--20}{Singapore}
\copyrightyear{2025}
\setcopyright{none}
\renewcommand\footnotetextcopyrightpermission[1]{}

\AtBeginDocument{%
  }

\begin{document}

\title{The conditional superiority of fast silicon sampling.}

\author{LAM Hock Yuen Nickolas}
\email{LA0001AS@e.ntu.edu.sg}
\affiliation{%
  \institution{Nanyang Technological University}
  \country{Singapore}
}

\author{VOO Ji Xuan}
\email{VOOJ0003@e.ntu.edu.sg}
\affiliation{%
  \institution{Nanyang Technological University}
  \country{Singapore}
}

\author{Xiangyu MA}
\email{xy.ma@ntu.edu.sg}
\affiliation{%
  \institution{Nanyang Technological University}
  \country{Singapore}
}

\begin{abstract}
\noindent Silicon sampling can produce surprisingly good population estimates at times. 
Does doing it fast attenuate such fidelity?
In this study, we extend and assess ongoing work in silicon sampling by comparing the algorithmic fidelity of ``fast'' and ``slow'' modes of silicon sampling among a nationally representative sample of Singaporean survey respondents.
We find that silicon sampling with contemporary frontier models remains a method in early development to be used only with great caution.
While silicon samples are able to produce moderately faithful estimates of population means, they continue to understate opinion variance and distort the latent contextual space behind human opinions.
Conditional on such limitations, we find ``fast'' modes of silicon sampling to be relatively superior to traditional ``slow'' modes of silicon sampling.
Fast silicon sampling is significantly more efficient in compute resources and run-time while being monotonically superior to slower modes of sampling in algorithmic fidelity.\newline 
 
\vspace{6pt}

\end{abstract}

\maketitle
\pagestyle{plain}

\pagebreak
\hypertarget{introduction}{%
\section{Introduction}\label{introduction}}

\begin{flushright}
  \begin{minipage}{0.5\linewidth}
    \itshape\raggedright
    Do the questions //
    answer themselves, //
    all wonder //
    brought to a reckoning?
    \par\medskip
    \upshape\raggedleft
    --- Robert Creeley, \emph{Fire}
    \par
  \end{minipage}
\end{flushright}
\bigskip

\noindent Large-language models can have surprisingly good intuitions about human thought and behavior.
Can we use these to create synthetic panels of survey and behavioral data, or simulate human opinion?
Attempts at doing just so have taken off in the last three years, most often under the banner of \emph{silicon sampling} (e.g. \citealt{argyle_out_2023}).
At their best, silicon samples can recover the opinion distributions of human subpopulations with surprising accuracy, reproducing not only population means but also some of the finer multivariate associations in social space that are inherent to human behavior (\citealt{argyle_out_2023,brand_using_2023, chen_emergence_2023}).
At their worst, they can be downright misleading, sharing only shallow similarities to human samples (\citealt{bisbee_synthetic_2024}).
But silicon sampling is a method in early and rapid development. 
What is true of silicon samples in 2023 does not necessarily follow with contemporary frontier models, given the enormous changes in model development and architecture.

In this paper, we evaluate the algorithmic fidelity of two distinct modes of silicon sampling -- fast and slow modes -- created by contemporary frontier models (OpenAI GPT 5.4).
We find that silicon sampling with contemporary frontier models remains a method in early development to be used only with great caution.
While silicon samples are able to produce moderately faithful estimates of population means, they continue to understate opinion variance and distort the latent contextual space behind human opinions.
Conditional on such limitations, we find ``fast'' modes of silicon sampling to be relatively superior to traditional ``slow'' modes of silicon sampling.
Fast silicon sampling is significantly more efficient in compute resources and run-time while being monotonically superior to slower modes of sampling in algorithmic fidelity.
 
Our work contributes to (1) ongoing methodological development of silicon sampling that is occurring across public opinion research and the social sciences, (2) improves evaluations of relational fidelity through geometric and Procrustes analysis, and (3) provides a deeper assessment of the algorithmic fidelity of silicon samples for an understudied national population. 
One, this paper contributes to ongoing work in the methodological development of silicon sampling by reviewing and evaluating the algorithmic fidelity of ``fast'' and ``slow'' modes of silicon sampling. 
We find fast silicon sampling to be conditionally superior to slow silicon sampling. 
Fast silicon sampling produces silicon samples that are as good as, if not better than, slow silicon sampling despite being considerably quicker and more efficient on compute.
However, the samples produced largely inherit the same set of weaknesses as slow silicon sampling; both diverge meaningfully from our reference human samples.
By convention, researchers tend to construct silicon samples \emph{slowly}.
They endow model sessions with distinct sociodemographic personas, and retrieve simulated responses to contextual prompts one API call at a time (Figure~\ref{fig:ss_process}).\footnote{Researchers rarely justify this choice explicitly, but adherence to precedence, technical limitations, and a desire to avoid question-order biases stand among the possible reasons why this preference exists presently. We elaborate more in Section~\ref{literature-review}.}
Improvements in frontier models, most notably the dramatic expansion in context windows and model performance, now make it possible to perform \emph{fast} modes of silicon sampling: in effect, we now construct elaborate prompts containing a large set of requests, eliciting a large set of synthetic tokens at a time.
Doing silicon sampling fast, rather than slow, saves both time and resources, if they are comparable.
They are -- conditional on accepting the limitations inherent to silicon sampling, fast sampling is superior to slow sampling.
Doing it fast does not attenuate or compromise algorithmic fidelity.

Second, we contribute to the evaluation of algorithmic fidelity by introducing techniques from geometric data analysis (\citealt{roux_multiple_2010}), a family of multivariate methods popular in the sociology of culture that is known for its emphasis on the relational properties between discrete categories in multivariate space.
Assessments of algorithmic fidelity have largely evaluated the ecological validity of population estimates item by item (e.g. \citealt{argyle_out_2023, santurkar_whose_2023, tan_can_2026}).
Investigations of multivariate fidelity have been limited to bivariate associations or the regression coefficients estimated on synthetic and human data (e.g. \citealt{bisbee_synthetic_2024, dominguez-olmedo_questioning_2024,shi_collapse_2026}).
Yet, neither item-level nor pairwise diagnostics can speak to the relational fidelity of the social space writ large.
Sociologists have long held that attitudes and tastes acquire their meaning from their position relative to other attitudes and tastes rather than from their marginal frequencies (e.g. \citealt{emirbayer_manifesto_1997, mohr_measuring_2020}).
A silicon sample could in principle reproduce every univariate mean in a survey while assembling those means into an opinion structure that no human population would recognize.
In this paper, we use multiple correspondence analysis (MCA) to assess such relational fidelity.
MCA, a technique most famously associated with the French sociologist Pierre Bourdieu (\citealt{bourdieu_distinction:_1984}), renders a battery of categorical items as a cloud of points in a low-dimensional space, in which two response categories lie close together to the extent that the same respondents tend to select both.
It has since become something of a workhorse for constructing spaces of taste and lifestyle in the sociology of culture (e.g. \citealt{meuleman_field_2013, cveticanin_art_2011, glevarec_structure_2021}), precisely because it treats the relations among categories, rather than the categories themselves, as the object of analysis.
Comparing two such spaces, however, is a recognized difficulty in that literature: cross-national and cross-sample applications of MCA typically establish similarity interpretively, by reading two maps side by side and observing that the same oppositions appear in both (e.g. \citealt{purhonen_methodological_2013, hanquinet_divergences_2025}).
We make that comparison explicit through Procrustes analysis, which superimposes one configuration onto another and returns scalar measures of their residual disagreement.
It is also discriminating.
Our silicon samples recover population means moderately well, yet the contextual spaces they generate bear little geometric resemblance to the configuration recovered from human respondents -- a divergence that item-level and pairwise checks leave entirely invisible.

Third, we provide a deeper assessment of the algorithmic fidelity of silicon samples for an understudied national population.
The empirical base for silicon sampling remains overwhelmingly American.
The foundational studies condition models on respondents drawn from Anglo-European surveys, and the critical literature that followed has largely audited those same instruments (e.g. \citealt{argyle_out_2023, bisbee_synthetic_2024, santurkar_whose_2023, dominguez-olmedo_questioning_2024}).
Fidelity established on American respondents may travel poorly, especially so for places like Singapore, a multilingual, multiethnic, and multireligious Chinese-majority city-state that departs in important ways from those very Anglo-European states (\citealt{pawar_survey_2025, liu_culturally_2025}).
Work on Singaporean silicon samples nonetheless remains scarce, and focused on the fidelity of population-level and subgroup estimates (e.g. \citealt{tan_can_2026, ku_silicon_2026}).
We complement this literature by further considering the fidelity of opinion dispersion, the associations among public opinions, and the contextual space these associations describe.

\hypertarget{literature-review}{%
\section{Literature review}\label{literature-review}}

Large language models (LLMs) are probabilistic machine learning models that are used to process and produce textual data. 
At a high level, LLMs resemble the autocomplete technologies that have become common on search engines and other digital spaces, but with considerably greater scale and sophistication (\citealt{bail_can_2024}). 
In this study, we extend ongoing work on the use of LLMs as synthetic surrogates to human respondents in public opinion surveys, a practice often referred to as \emph{silicon sampling}. 
The intuition behind silicon sampling is simple. 
Homo silicus has advanced to such a state that many LLMs are able to converse and reason in ways that are virtually indistinguishable from humans. 
If this were so, then it may be possible to productively sample responses from LLMs, the same way we would from human populations. 
Following this intuition, Argyle et al. (\citeyear{argyle_out_2023}) introduced the idea of silicon sampling, arguing that it may be possible to instead think of them in productive terms as a model of the ``complex reflection of the many various patterns of association between ideas, attitudes, and contexts present among humans.''
The algorithmic fidelity of LLMs, when appropriately conditioned, means that researchers may be able to induce the model to produce outputs that correlate with the attitudes, opinions, and experiences of distinct human subpopulations. 
In the years since, this methodological turn has attracted substantial scholarly and commercial attention, prompting both further development and critique (Figure~\ref{fig:slow_fast_cumulative}).
Work is also underway exploring the applications of synthetic surrogates of human respondents in domains beyond survey research, from agent-based modeling to psychological experiments, behavioral economics experiments, market research, and philosophical reasoning (\citealt{chen_emergence_2023,dillion_can_2023,horton_large_2023, park_generative_2023, goli_frontiers_2024, shi_collapse_2026}). 

In this study, we apply silicon sampling techniques to a public opinion survey fielded to a nationally representative sample of Singaporeans.
There is limited work on the algorithmic fidelity of silicon samples of Singaporean public opinion.
Much of the AI/ML research from Singapore has hitherto been centered around non-Singaporean contexts (\citealt{ku_silicon_2026}), drawing on surveys such as the Survey of Public Participation in the Arts (SPPA) which are fielded on nationally representative samples from the United States (e.g. \citealt{ma_not-quite-human_2026}). 
To date, we have only identified one study that explicitly focused on conducting silicon sampling in the specific context of Singaporean public opinion (\citealt{tan_can_2026}).
The paper focused on understanding the fidelity of silicon sampling when applied to different subgroup populations -- for instance, they compared the accuracy of various
subgroups in Singapore such as Chinese Males and Indian Muslims (\citealt{tan_can_2026}). 
They find silicon sampling to be more accurate for ``young, male, Chinese, and Christian'' personas (\citealt[p.24571]{tan_can_2026}); in contrast, it is much less accurate for the demographic of Malay respondents aged 35-44. 

\subsection{Fast and slow forms of silicon sampling.}
In particular, we discuss and compare the algorithmic fidelity of two modes of silicon sampling, which we stylize as slow silicon sampling and fast silicon sampling.
\emph{Slow silicon sampling} refers to repeated zero- or one-shot prompting of large language models with the intention of eliciting a small set of synthetic responses from the model. 
Each prompt contains a constrained and well-defined request to act as a human surrogate, and is mapped to a unitary response.
The conventional approaches that silicon sampling researchers have followed hew to the model of slow silicon sampling. 
Researchers using silicon sampling usually generate synthetic responses to survey questions one at a time, proceeding procedurally through a questionnaire while resetting the model after each set of input prompts (e.g. \citealt{argyle_out_2023,bisbee_synthetic_2024}).
\emph{Fast silicon sampling}, on the other hand, refers to the elaborate zero- or one-shot prompting of large language models with the intention of eliciting a large set of synthetic responses from the model.
Under fast sampling paradigms, a researcher would issue a singular complex prompt that contains a large set of requests; this singular prompt should elicit a large set of synthetic responses, albeit in a single batch of output tokens.
Researchers using fast silicon sampling may simulate entire sets of psychological experiments at a time (e.g. \citealt{almeida_exploring_2023, park_generative_2023}).

\begin{figure}
  \centering
  \caption{Cumulative use of slow and fast silicon sampling in preprints over time}
  \label{fig:slow_fast_cumulative}
  \includegraphics[width=\linewidth]{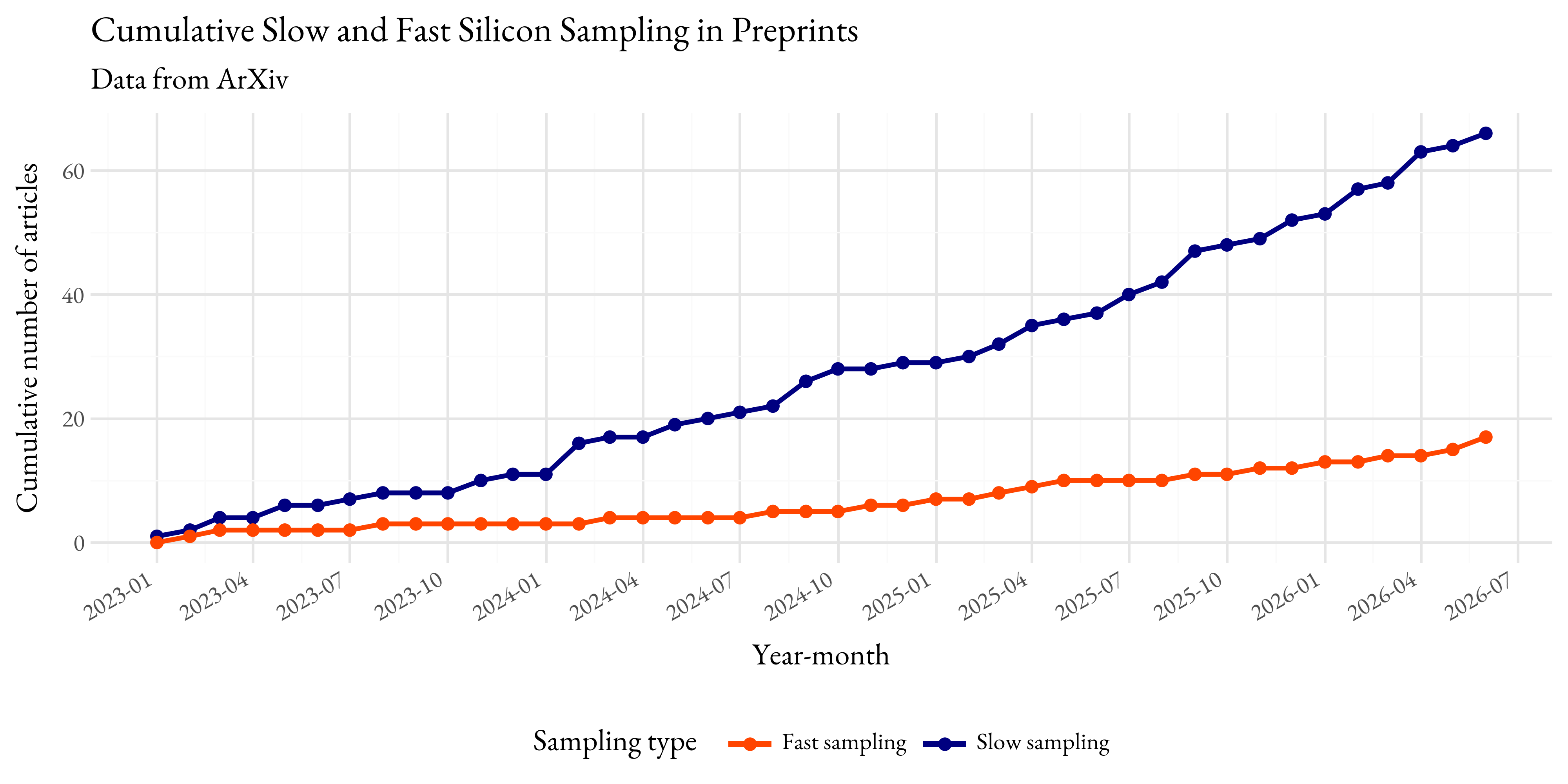}

  \vspace{2pt}
  \begin{minipage}{\linewidth}
    \footnotesize
    \emph{Note.} Preprints labelled as using both modes are counted in both series. The final month covers a partial year. Source: \emph{ArXiv}.
  \end{minipage}
\end{figure}

To assess researchers' usage of slow and fast silicon sampling, we reviewed and hand-coded a corpus of preprints on \emph{ArXiv} that perform silicon sampling in a substantive way. We instructed an AI agent (Claude Code) to search \emph{ArXiv} from 1/1/2023 to 7/1/2026 for all papers performing silicon sampling, defined broadly as:

\begin{quote}
\itshape ``...using LLMs or AI models to produce synthetic surrogates of human respondents. This could be in the context of a survey, or in the context of a lab experiment... Do a broad search for related wildcards as well — this could be silicon sampling, silicon surrogates, silicon/synthetic twins and so on.''
\end{quote}

\noindent The search was intentionally broad rather than restricted to a single keyword, in order to capture (a) state-of-the-art techniques for LLM-based respondent simulation and (b) critical analyses and methodological critiques of silicon sampling, rather than only papers that used the term ``silicon sampling'' itself. 
Search terms therefore included direct variants of the phrase (``silicon sampling,'' ``silicon surrogates,'' ``silicon twins,'' ``synthetic twins''), as well as adjacent framings used in the literature (e.g., algorithmic fidelity, LLM-based social/economic simulation, persona-conditioned survey response generation).
A complementary systematic crawl of Google Scholar, searched year by year through a June 2026 cutoff, was separately commissioned to cross-check coverage against the ArXiv-only search.
We identified 98 papers from our search, of which 75 engage substantively in silicon sampling.

Over this period (1/1/2023 to 6/1/2026), we find slow sampling to have clearly been the more preferred mode of silicon sampling among \emph{ArXiv} preprints (Figure~\ref{fig:slow_fast_cumulative}).
Of the 75 preprints, 66 use slow sampling, compared with 17 that use fast sampling; eight use both modes and therefore appear in both series.
While few if any studies offer explicit justifications for why they favor slow sampling over fast sampling, we infer three general reasons as to why.
One, it follows established precedent. 
A large share of papers adopting slow silicon sampling follow Argyle et al. (\citeyear{argyle_out_2023})'s original sampling protocol (e.g. \citealt{sun_random_2024,cummins_threat_2025, heyde_vox_2026}), with many making direct references to these inherited procedures.
Two, they are necessary because of technical limitations.
Large-language models have developed significantly over the past few years, and one area of improvement most significant for silicon sampling has been the expansion of context windows to accommodate large input prompts that would previously have been impossible.
Many preprints also mentioned technical limitations in passing when describing their silicon sampling procedures.
For example, preprints mention the degradation of model performance when given long prompts (e.g. \citealt{lee_can_2024}); others keep prompts short to keep output robust to poor model performance (e.g. \citealt{griffin_susceptibility_2023}).
Three, they avoid question-order biases.
Many preprints mention that they engage in repeated prompting in fresh and isolated LLM sessions to preclude any memory or carry-over of context (e.g. \citealt{griffin_susceptibility_2023}).

The case for fast silicon sampling is made primarily on the grounds of efficiency.
Fast sampling uses fewer tokens, over fewer API calls, to produce the same set of output tokens. 
It is significantly less taxing on compute; silicon samples can be assembled with fast sampling procedures far more cheaply and quickly (see Table~\ref{tab:sampling_compute} for a comparison from our study).
Preprints that have adopted fast sampling procedures frequently cite cost and scalability as the reasons why (e.g. \citealt{zhou_chatgpt_2025,tzachristas_guided_2026,miklian_stochastic_2026}).
Ironically, fast sampling is also closer to how humans procedurally answer surveys. 
Human respondents cannot reset caches and memories; indeed, question-ordering biases remain a pain-point that vexes survey researchers, who in turn have developed a variety of approaches to control or accommodate them (\citealt{schuman_questions_1981}).

Fast sampling might be efficient -- but are the silicon samples produced thereof any good?
To our knowledge, there is no work providing such assessments.
Ultimately, the attractiveness of fast sampling is grounded in its fidelity relative to slow sampling.
There is little purpose to fast silicon sampling if the synthetic samples produced are worse or less faithful to real population data.
But if they are close to -- or \emph{better than} -- those from slow samples, then anyone producing silicon samples ought to seriously consider switching, given its superior efficiency.
In what follows, we provide exactly such a comparison between the two.

\hypertarget{data-and-methods}{%
\section{Data and methods}\label{data-and-methods}}

\hypertarget{data-source}{%
\subsection{Data source}\label{data-source}}

We draw on primary data from the Singapore sample ($n = 2,012$) of the World Values Survey (WVS) Wave 7, conducted between 2017 and 2022 (\citealt{haerpfer_world_2022}). 
Data from the WVS serve as the human reference points upon which we build our silicon samples.
The WVS is a repeated cross-sectional comparative survey of public opinion on religion, morality, values, and politics.
First fielded in 1981, it has grown into one of the largest non-commercial cross-national studies of human beliefs and values, spanning more than 120 societies across seven waves, and is widely used both in social scientific research (e.g. \citealt{inglehart_modernization_2000}) and as a human-opinion benchmark in computer science (e.g. \citealt{santurkar_whose_2023}).
Interviews are primarily conducted face-to-face in respondents' homes (\citealt{haerpfer_world_2022}).
Each national survey, including Singapore's, targets a probability sample of the adult resident population including both nationals and residents, unless migrants exceed half the population.
Summary statistics on the WVS respondents included in this study can be found in Table \ref{tab:wvs_covariate_summary}.

\subsubsection{Key measures} 

We focus on two batteries of public opinion items from the WVS, eight items concerning the perceived effects of immigration and 19 items concerning ethical norms. 
For each immigration item, respondents indicate whether they disagree (0), find it hard to say (1), or agree (2). 
All eight measures run from 0 to 2, with higher values consistently indicating a more favorable view of immigration.\footnote{To make it so, we reverse-code four items phrased as adverse consequences -- increased crime, increased terrorism risk, increased unemployment, and social conflict -- while leaving the four positively phrased items unchanged.} 
The ethical norm items ask whether each behavior or practice is justifiable, using a ten-point scale from 1 (\emph{never justifiable}) to 10 (\emph{always justifiable}). 
We retain this coding, so higher values indicate greater acceptance of the behavior or practice. 
Summary statistics of these public opinion items can be found in Table~\ref{tab:wvs_attitude_summary}.
 
\hypertarget{constructing-silicon-samples}{%
\subsection{Constructing silicon samples}\label{constructing-silicon-samples}}

\begin{figure}
  \centering
  \caption{Process of slow and fast silicon sampling}
  \label{fig:ss_process}
  \fbox{\includegraphics[width=0.95\textwidth]{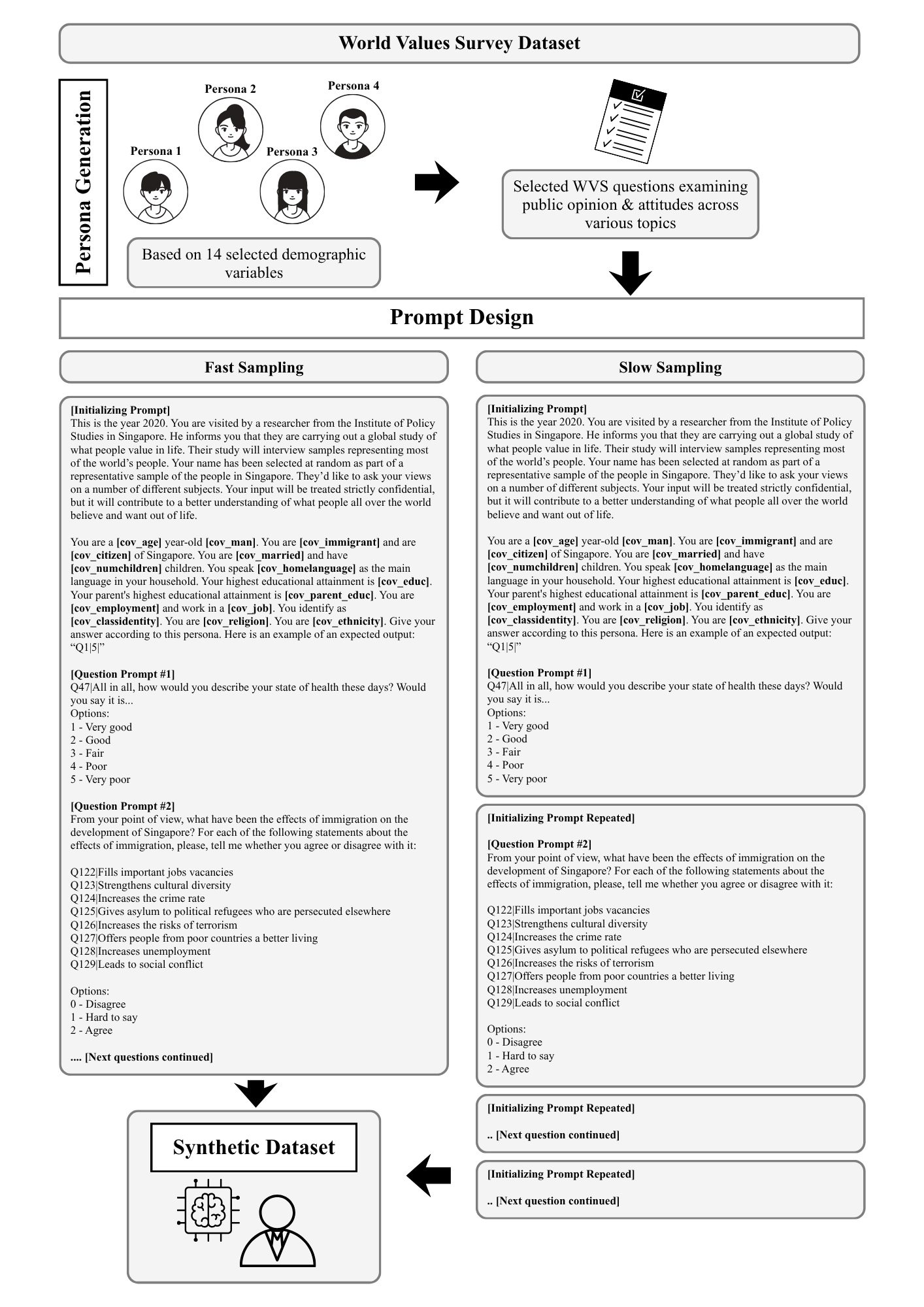}}
\end{figure} 

We construct silicon samples using two distinct data-generation pipelines, first using a conventional ``slow'' silicon sampling approach, then a second ``fast'' silicon sampling approach (Figure~\ref{fig:ss_process}).
We use a frontier model from OpenAI, GPT-5.4, with weights from March 5th, 2026.
We set maximum output tokens to $2,500$, and set temperature to $1.0$.
For each respondent $i$ in the WVS sample, we generate 30 silicon surrogates using each of the two modes of silicon sampling.
In total, this gives us a total of $30 \times 2,012$ respondents in each of the silicon samples.
Table~\ref{tab:sampling_compute} summarizes the computational costs of the two approaches. 
Generating the complete response set of a single respondent under slow sampling requires approximately 20 seconds and 4,000 tokens, compared with 3 seconds and 1,450 tokens under fast sampling.
This makes fast sampling about 670 percent faster while using 64 percent fewer tokens per run. 
Across the complete set of 60,360 synthetic responses, this accumulates to an estimated aggregate runtime of 335.3 hours for slow sampling and 50.3 hours for fast sampling, and a corresponding token consumption of 241.4 million and 87.5 million respectively. 
Table~\ref{tab:sampling_compute} also provides the estimated costs from the frontier model provider (OpenAI) at the time of writing.

Both slow and fast sampling proceed as follows, with (1) a system prompt along with an initializing (``one-shot'') prompt, (2) the question prompt, and (3) the retrieval of LLM-output. 
They differ only in the size of the question prompt and output retrieved with each API call.
Slow sampling makes repeated API calls with small question prompts, collecting a slice of desired final output each time; fast sampling makes a unitary API call and collects the desired final output all at once.

\begin{table*}
  \centering
  \caption{Costs of slow and fast silicon sampling}
  \label{tab:sampling_compute}
  \small
  \setlength{\tabcolsep}{8pt}
  \begin{tabular}{lrrrrrr}
    \toprule
    & \multicolumn{2}{c}{Runtime} & \multicolumn{2}{c}{Tokens} & \multicolumn{2}{c}{Cost (USD)} \\
    \cmidrule(lr){2-3}\cmidrule(lr){4-5}\cmidrule(lr){6-7}
     & Per run (s) & Aggregate (hr) & Per run & Aggregate (M) & Per run & Aggregate \\
    \midrule
    Slow Sampling & 20 & 335.3 & 4{,}000 & 241.440 & \$0.010 & \$603.60 \\
    Fast Sampling & 3 & 50.3 & 1{,}450 & 87.522 & \$0.005 & \$301.80 \\
    \bottomrule
  \end{tabular}
\end{table*}

\subsubsection{Design of initializing prompt}

We preface each API call with an initializing prompt that provides the LLM with the necessary context for the task, along with a ``one-shot'' example of the desired output.
To ensure that our prompt design accurately reflected the original administration procedure of the WVS, we began each prompt with an introductory passage adapted from the WVS interviewer script, reflecting the general survey purpose, response context, and instructions presented to WVS human respondents. 

\begin{quote}
  \small
  This is the year 2020. You are visited by a researcher from the Institute of Policy Studies in Singapore. He informs you that they are carrying out a global study of what people value in life. Their study will interview samples representing most of the world's people. Your name has been selected at random as part of a representative sample of the people in Singapore. They'd like to ask your views on a number of different subjects. Your input will be treated strictly confidential, but it will contribute to a better understanding of what people all over the world believe and want out of life.
  \end{quote}

\noindent Then, for each respondent in the WVS sample, we construct and append a demographic persona using a combination of 14 demographic variables. The demographic persona was subsequently prompted through the following format:

\begin{quote}
\small 
\noindent You are a [AGE] year-old [GENDER]. You are [IMMIGRANT\_STATUS] and are [CITIZEN\_STATUS] of Singapore. You are [MARRIED] and have [NUMCHILDREN] children. You speak [HOMELANGUAGE] as the main language in your household. Your highest educational attainment is [EDUC]. Your parents' highest educational attainment is [PARENT\_EDUC]. You are [EMPLOYMENT\_STATUS] and work in a [JOB]. You identify as [CLASSIDENTITY]. You are [RELIGION]. You are [ETHNICITY]. Give your answer according to this persona.
\medskip
\noindent Here is an example of an expected output: ``Q1|5''
\par
\end{quote}

\subsubsection{Constructing the question prompt}

The question prompt is where slow and fast sampling depart from one another (Figure~\ref{fig:ss_process}). 
Slow sampling involves the submission of repeated, non-cached inputs to a large-language model until the desired output set is assembled.
When performing slow sampling, we take a segmented prompting approach in which we decompose a survey questionnaire into atomic constituent units. 
These units may comprise a singular question, or a small bloc of questions. 
Only one unit of questions is appended to the initializing prompt and submitted to a large-language model through an API call.
After each API call, the session cache is reset before another set of prompts is submitted to a fresh session.
While allowing the model to focus on narrower individual subsets of survey items at a time, this strategy also required significantly greater investments of computational time and cost. 
Repeated inclusions of the survey introduction and demographic personas over repeated model calls also corresponded to a greater number of tokens required for each model prompt.

On the other hand, fast sampling involves the submission of a singular elaborate prompt that returns the desired output set all at once.
When performing fast sampling, we construct a single large prompt that includes all survey items.
This large prompt is then appended to the initializing prompt before being submitted to the model in a single API call.
Due to only requiring a single model call for each respondent, this strategy greatly increases efficiency of silicon sample generation in terms of both time and cost.

\subsubsection{Retrieving model output}
Outputs from the large-language models are then saved into a .json file.
For slow sampling, we process and save survey responses on the respondent-iteration level, i.e. after a complete set of repeated API queries is made.
For fast sampling, we process and save survey responses after every API query, since each API query produces a complete set of output.

\hypertarget{estimands}{%
\subsection{Estimands}\label{estimands}}

We compare three ecological estimates: (A1) the population mean of a public opinion among a given silicon sample, (A2) the mean Jaccard distance between a WVS respondent and their silicon surrogate in a given silicon sample, and (A3) the population variance of a public opinion among a given silicon sample.
\subsubsection{Mean, variance, and distance.}

To compare the population mean of public opinions, we compute the sample mean for each attitude item $k$ in each silicon sample, $\bar{x}_k$, and compare it against the corresponding WVS mean, $\bar{x}^{\text{WVS}}_k$, reporting the raw difference $\Delta_k = \bar{x}_k - \bar{x}^{\text{WVS}}_k$.
Raw deltas are not comparable across the two domains, whose items run on different scales ($0$ to $2$ for immigration, $1$ to $10$ for morals), so we also report a standardized bias, $\Delta_k / s^{\text{WVS}}_k$, dividing each delta by that item's WVS standard deviation.\footnote{We report this in place of significance stars because at $n=60{,}360$ per domain, a two-sample test rejects the null for all but the smallest deltas.}

To compare the distance between silicon samples and the WVS data, we use a distance measure based on Jaccard similarity (``Jaccard distance'') to measure the differences between silicon and WVS attitudes.
Jaccard similarity is a statistical measure used to quantify how similar two sets are by comparing their intersection to their union. 
Let $t_{i}$ denote the set of item--response pairs recorded for WVS respondent $i$, and $\tilde{t}_{i}$ the corresponding set for their silicon counterpart.
Since both answer the same battery of $K$ items, their overlap
\[J(t_{i}, \tilde{t}_{i}) = \frac{|t_{i} \cap \tilde{t}_{i}|}{K}\]

\noindent is the share of items on which the silicon respondent gives exactly the answer the real respondent gave.
We define the Jaccard distance as $D_{i} = 1 - J(t_{i}, \tilde{t}_{i})$, bounded below by $0$ for a perfectly reproduced respondent and above by $1$ for one who agrees on no item at all, and average it over all 2,012 respondents and 30 resampling iterations within each domain.

To compare the population variance of public opinions, we compare each attitude item $k$'s standard deviation in the silicon sample, $s_k$, against $s^{\text{WVS}}_k$.
We test for equality of variances using Fligner-Killeen tests, which make no normality assumption and are robust to the heavily skewed, floor-bounded distributions that many of these items exhibit.

\subsubsection{Bivariate associations}
We consider how well silicon samples preserve the relationality of public opinions by measuring the bivariate associations among public opinions within each domain.
Because these public opinion items are categorical, we measure association using Cramer's V rather than a product-moment correlation.
For two items cross-tabulated into an $r \times k$ contingency table over $n$ responses,
\[V = \sqrt{\frac{\chi^{2}}{n \, (\min(r,k) - 1)}},\]
where $\chi^{2}$ is the usual Pearson statistic for that table.

We compute $V$ for every unique pair of items within a domain, pooling respondents across all 30 resampling iterations.
Items with no response variation must be dropped, since $V$ is undefined when either variable is constant, which leaves 21 pairs for immigration and 78 for morals.
This yields a full matrix of pairwise associations per source and domain, which we compare against the WVS matrix in two complementary ways.
One, we compute the Pearson correlation between a silicon sample's pairwise associations and the corresponding WVS values.
Two, we compute the mean signed deviation, the average of $V - V^{\text{WVS}}$ across pairs, which asks whether it gets the \emph{level} right: negative values indicate that the silicon sample understates associations, positive values that it overstates them.
For both we include a same-population bootstrap resample of the WVS as a benchmark, which indicates how much apparent infidelity is attributable to sampling noise alone.

\subsubsection{Contextual space}

We use multiple correspondence analysis (MCA) to transform public opinions into a contextual space that captures the contextual relations among the opinions.
Then, we use Procrustes statistics to compare contextual spaces across silicon samples and WVS data.

Multiple correspondence analysis (MCA), the categorical analogue of principal components analysis, transforms an indicator matrix of item-responses in public opinions into a cloud of points in a low-dimensional space (\citealt{roux_multiple_2010}).
Every response category receives a coordinate, and two categories sit close together to the extent that the same respondents tend to select both, so the fitted configuration renders the structure of opinion as a shape that can be inspected and compared.
We fit an MCA separately to each source within each domain, retaining the first two dimensions and fixing the random seed so the solutions are reproducible.\footnote{We use the MCA implementation from the \emph{prince} package in Python.}
Explained inertia is reported using Greenacre's correction, since indicator-matrix coding inflates total inertia with within-variable blocks and consequently understates the variance the leading dimensions actually account for (\citealt{nenadic_correspondence_2007}).

To compare the MCA configurations, we perform Procrustes analysis, a common technique in multivariate analysis which finds the isotropic dilation, translation, reflection, and rotation that best matches one configuration to another until the sum of squared differences between them is minimized (\citealt{cox_multidimensional_2008, krzanowski_principles_2023, borg_mds_2005}).
We hold the WVS configuration fixed as the reference and align each other configuration onto it, matching points on the response categories the two configurations share.
We summarize the differences between configurations with two goodness-of-fit statistics, (a) Gower's $M^{2}$ and (b) Tucker's coefficient of congruence $\varphi$.
Let $X$ be the $p \times 2$ matrix of WVS coordinates and $Y$ the corresponding matrix from the sample being compared, over the $p$ response categories the two configurations share.
We first center both column-wise and normalize each to unit Frobenius norm,
\[\tilde{X} = \frac{X - \mathbf{1}\bar{x}^{\top}}{\lVert X - \mathbf{1}\bar{x}^{\top} \rVert_{F}}, \qquad \tilde{Y} = \frac{Y - \mathbf{1}\bar{y}^{\top}}{\lVert Y - \mathbf{1}\bar{y}^{\top} \rVert_{F}},\]
so that what remains to be compared is shape alone. 
Gower's $M^{2}$ is then the minimized residual sum of squares over rotations $R$ (orthogonal, so reflections are admitted) and dilations $s > 0$,
\[M^{2} = \min_{R, \, s} \, \bigl\lVert \tilde{X} - s\,\tilde{Y}R \bigr\rVert_{F}^{2} = 1 - \Bigl(\textstyle\sum_{i} \sigma_{i}\Bigr)^{2},\]
where $\sigma_{1}, \sigma_{2}$ are the singular values of $\tilde{Y}^{\top}\tilde{X}$ (\citealt{gower_generalized_1975}).
Because both configurations are normalized before fitting, $M^{2}$ is symmetric in its two arguments and bounded on $[0,1]$, with $0$ for identical shapes and $1$ for no recoverable correspondence at all.\footnote{We compute $M^{2}$ using the implementation from the \emph{scipy} package in Python.}

$M^{2}$ scores the configuration as a whole, which leaves open whether a poor fit is spread evenly through the space or concentrated in a single axis.
To address this, we also compute Tucker's coefficient of congruence, which compares the two configurations one dimension at a time.
Let $\hat{Y} = s\,\tilde{Y}R$ for the aligned configuration at the optimal $R$ and $s$, and indexing the dimensions by $j$,
\[\varphi_{j} = \frac{\sum_{i} \tilde{x}_{ij}\,\hat{y}_{ij}}{\sqrt{\sum_{i} \tilde{x}_{ij}^{2} \; \sum_{i} \hat{y}_{ij}^{2}}},\]
the cosine between corresponding columns of the reference and the aligned configuration, taken on centered coordinates so that a shared centroid offset cannot inflate it (\citealt{tucker_method_1951}).
It equals $1$ when dimension $j$ is reproduced exactly up to scale and $0$ when the two orderings along that axis are unrelated.

\hypertarget{results}{%
\section{Results}\label{results}}

\subsection{How faithful are silicon estimates of population means?}\label{sec:fidelity-means}

  \begin{table}[htbp]
  \caption{Comparison of Means from Slow and Fast Silicon Sampling}
  \label{tab:means_comparison}

  \begin{tabular*}{\linewidth}{@{\extracolsep{\fill}}p{0.4\textwidth}llrlr}
  \toprule
    & \multicolumn{1}{c}{WVS} & \multicolumn{2}{c}{Slow sampling} & \multicolumn{2}{c}{Fast sampling} \\
  \cmidrule(lr){2-2} \cmidrule(lr){3-4} \cmidrule(lr){5-6}
    & Mean & Mean & Delta & Mean & Delta \\
  \midrule\addlinespace[2.5pt]
  \multicolumn{6}{l}{\textbf{Immigration}} \\
  \varindent Fills impt. job vacancies & 1.24 & 2.00 & 0.76 [0.87] & 2.00 & 0.75 [0.86] \\
  \varindent Strengthens cultural diversity & 1.20 & 1.94 & 0.74 [0.85] & 1.91 & 0.72 [0.83] \\
  \varindent Decreases crime & 1.04 & 1.51 & 0.47 [0.55] & 1.08 & 0.04 [0.05] \\
  \varindent Provides asylum to the persecuted & 0.52 & 1.76 & 1.24 [1.51] & 1.30 & 0.79 [0.96] \\
  \varindent Decreases terrorism risk & 1.10 & 1.23 & 0.14 [0.16] & 1.01 & -0.08 [-0.09] \\
  \varindent Gives the poor a better living & 1.48 & 2.00 & 0.52 [0.66] & 2.00 & 0.52 [0.66] \\
  \varindent Decreases unemployment & 0.84 & 0.96 & 0.12 [0.14] & 0.63 & -0.22 [-0.25] \\
  \varindent Decreases social conflict & 0.89 & 0.91 & 0.03 [0.03] & 0.91 & 0.02 [0.02] \\
  \midrule\addlinespace[2.5pt]
  \multicolumn{6}{l}{\textbf{Ethical norms}} \\[2.5pt]
  \varindent Benefits fraud & 2.51 & 1.00 & -1.51 [-0.69] & 1.00 & -1.51 [-0.69] \\
  \varindent Fare evasion on public transport & 1.80 & 1.44 & -0.36 [-0.22] & 1.03 & -0.77 [-0.46] \\
  \varindent Theft & 1.33 & 1.00 & -0.33 [-0.30] & 1.00 & -0.33 [-0.30] \\
  \varindent Tax evasion & 1.51 & 1.01 & -0.50 [-0.37] & 1.00 & -0.51 [-0.38] \\
  \varindent Taking bribes & 1.39 & 1.00 & -0.39 [-0.33] & 1.00 & -0.39 [-0.33] \\
  \varindent Homosexuality & 3.38 & 5.05 & 1.66 [0.56] & 4.20 & 0.82 [0.27] \\
  \varindent Prostitution & 2.73 & 3.08 & 0.36 [0.15] & 2.55 & -0.18 [-0.07] \\
  \varindent Abortion & 3.22 & 3.80 & 0.58 [0.21] & 3.41 & 0.19 [0.07] \\
  \varindent Divorce & 4.14 & 6.02 & 1.88 [0.66] & 6.24 & 2.10 [0.73] \\
  \varindent Fornication & 3.96 & 4.63 & 0.67 [0.22] & 4.69 & 0.73 [0.24] \\
  \varindent Suicide & 2.42 & 1.77 & -0.65 [-0.29] & 1.42 & -1.00 [-0.44] \\
  \varindent Euthanasia & 4.06 & 4.34 & 0.28 [0.09] & 4.10 & 0.05 [0.02] \\
  \varindent A man beating his wife & 1.30 & 1.00 & -0.30 [-0.27] & 1.00 & -0.30 [-0.27] \\
  \varindent Parents beating children & 3.33 & 2.94 & -0.39 [-0.16] & 2.43 & -0.89 [-0.36] \\
  \varindent Violence against other people & 1.57 & 1.00 & -0.57 [-0.42] & 1.00 & -0.57 [-0.42] \\
  \varindent Terrorism & 1.29 & 1.00 & -0.29 [-0.26] & 1.00 & -0.29 [-0.26] \\
  \varindent Casual sex & 2.81 & 3.18 & 0.37 [0.14] & 3.74 & 0.93 [0.36] \\
  \varindent Political violence & 1.53 & 1.00 & -0.53 [-0.40] & 1.00 & -0.53 [-0.40] \\
  \varindent Death penalty & 4.70 & 5.39 & 0.69 [0.23] & 5.33 & 0.63 [0.21] \\
  \bottomrule
  \end{tabular*}

  \vspace{2pt}
  {\footnotesize \raggedright \setlength{\baselineskip}{9pt} \emph{Note.} Delta is the difference between the silicon sample mean and the WVS mean. Bracketed values are the standardized bias (delta divided by the WVS item SD from Table~\ref{tab:wvs_attitude_summary}).\par}

  \end{table}
  
\subsubsection{Ecological estimates of population means from silicon samples are moderately faithful.}
Averaged across all 27 items, the mean standardized bias is 0.40 SD for slow sampling and 0.37 SD for fast sampling, estimates that are typically considered moderately similar by conventional benchmarks (\citealt{cohen_statistical_2013,cohen_power_1992}).
Ecological estimates of means are better for public opinions on ethical norms than on immigration.
Immigration attitudes carry roughly twice the standardized bias of moral attitudes (slow: 0.60 SD immigration vs.\ 0.31 SD moral; fast: 0.47 SD vs.\ 0.33 SD); immigration items also carry the largest synthetic bias (fills important job vacancies; strengthens cultural diversity; and provides asylum).
This item-level picture is corroborated by a more holistic, respondent-level distance measure reported in Table~\ref{tab:jdist_comparison}.
Silicon respondents' answer patterns overlap with their real-world counterpart's only around 40 percent of the time. 
Averaged across 2,012 respondents and 30 resampling iterations, this whole-profile distance sits at 0.59--0.60 out of a maximum of 1 for both domains and both sampling strategies. 

\begin{table}[htbp]
  \caption{Jaccard Distance Between Silicon Sample Respondents and Their Matched WVS Respondent}
  \label{tab:jdist_comparison}

  \begin{tabular*}{\linewidth}{@{\extracolsep{\fill}}lccccc}
  \toprule
   & \multicolumn{2}{c}{Slow sampling} & \multicolumn{2}{c}{Fast sampling} & \\
  \cmidrule(lr){2-3} \cmidrule(lr){4-5}
   & Mean & SD & Mean & SD & $\Delta$ (Slow $-$ Fast) \\
  \midrule
  Immigration attitudes & 0.602 & 0.183 & 0.590 & 0.188 & $0.012$*** \\
  Moral attitudes & 0.589 & 0.142 & 0.569 & 0.149 & $0.020$*** \\
  \bottomrule
  \end{tabular*}

  \vspace{2pt}
  {\footnotesize \raggedright \setlength{\baselineskip}{9pt} \emph{Note.} Jaccard distance is computed between each silicon respondent's full multi-item response vector and their matched WVS respondent, averaged over 2,012 respondents $\times$ 30 resampling iterations ($N=60{,}360$ per domain). $\Delta$ is the slow-sampling mean minus the fast-sampling mean; significance is from Welch's two-sample $t$-tests comparing fast to slow sampling within each domain (immigration: $t=11.9$; morals: $t=23.6$): * $p<0.05$, ** $p<0.01$, *** $p<0.001$.\par}

  \end{table}

\subsubsection{Fast sampling is no less faithful than slow sampling.}
More surprisingly, we find that this moderate, domain-dependent fidelity is indifferent to the mode of silicon sampling. 
Averaged across all 27 items, mean absolute delta from WVS is 0.605 for slow sampling and 0.587 for fast sampling; in standardized terms, 0.397 SD vs.\ 0.371 SD. 
This agreement goes beyond magnitude to direction. 
In 24 of 27 items (89\%), slow and fast sampling err on the \emph{same side} of the WVS mean, even where the size of the miss differs substantially (e.g.\ divorce, casual sex). 
Where the two strategies do diverge, the direction of the advantage is domain-specific rather than universal. 
Fast sampling carries a real, fairly systematic edge for public opinions on immigration attitudes, but for opinions on ethical norms the two modes of sampling are difficult to split.
On immigration items, fast sampling is closer to the WVS mean on 6 of 8 items -- most dramatically on ``decreases crime'' (delta $+0.04$ fast vs.\ $+0.47$ slow) and ``provides asylum to the persecuted'' ($+0.79$ vs.\ $+1.24$) -- with slow only closer on ``decreases unemployment.'' 
On ethical norms items, however, there is no consistent winner. 
Fast sampling is closer on homosexuality ($+0.82$ vs.\ $+1.66$), prostitution, abortion, and euthanasia, while slow is closer on suicide ($-0.65$ vs.\ $-1.00$), parents beating children ($-0.39$ vs.\ $-0.89$), casual sex ($+0.37$ vs.\ $+0.93$), and divorce --- splitting roughly 7 slow-wins to 5 fast-wins across the 12 non-floor moral items. 

These observations are corroborated by comparisons of Jaccard distances between silicon samples and WVS data (Table~\ref{tab:jdist_comparison}).
Slow and fast sampling land within hundredths of one another: mean Jaccard distance 0.602 vs.\ 0.590 for immigration, 0.589 vs.\ 0.569 for morals.
Further, the gap surprisingly favors fast sampling on both domains, with fast sampling being 0.012 and 0.020 closer in Jaccard distance to WVS data than slow sampling. 

\subsection{How faithful are silicon estimates of population variance?}\label{sec:fidelity-variance}

  \begin{table}[htbp]
  \caption{Comparison of Standard Deviations from Slow and Fast Silicon Sampling}
  \label{tab:sd_comparison}

  \begin{tabular*}{\linewidth}{@{\extracolsep{\fill}}p{0.5\textwidth}lllll}
  \toprule
    & \multicolumn{1}{c}{WVS} & \multicolumn{2}{c}{Slow sampling} & \multicolumn{2}{c}{Fast sampling} \\
  \cmidrule(lr){2-2} \cmidrule(lr){3-4} \cmidrule(lr){5-6}
    & SD & SD & Delta & SD & Delta \\
  \midrule\addlinespace[2.5pt]
  \multicolumn{6}{l}{\textbf{Immigration}} \\
  \varindent Fills impt. job vacancies & 0.87 & 0.01 & -0.86*** & 0.03 & -0.84*** \\
  \varindent Strengthens cultural diversity & 0.87 & 0.21 & -0.66*** & 0.25 & -0.62*** \\
  \varindent Decreases crime & 0.86 & 0.51 & -0.35*** & 0.41 & -0.45*** \\
  \varindent Provides asylum to the persecuted & 0.82 & 0.31 & -0.51*** & 0.38 & -0.44*** \\
  \varindent Decreases terrorism risk & 0.88 & 0.32 & -0.56*** & 0.25 & -0.63*** \\
  \varindent Gives the poor a better living & 0.79 & 0.00 & -0.79*** & 0.00 & -0.79*** \\
  \varindent Decreases unemployment & 0.88 & 0.42 & -0.46*** & 0.51 & -0.37*** \\
  \varindent Decreases social conflict & 0.87 & 0.23 & -0.64*** & 0.24 & -0.63*** \\
  \midrule\addlinespace[2.5pt]
  \multicolumn{6}{l}{\textbf{Ethical norms}} \\[2.5pt]
  \varindent Benefits fraud & 2.20 & 0.00 & -2.20*** & 0.00 & -2.20*** \\
  \varindent Fare evasion on public transport & 1.66 & 0.38 & -1.28*** & 0.12 & -1.54*** \\
  \varindent Theft & 1.11 & 0.00 & -1.11*** & 0.00 & -1.11*** \\
  \varindent Tax evasion & 1.34 & 0.05 & -1.29*** & 0.02 & -1.32*** \\
  \varindent Taking bribes & 1.19 & 0.00 & -1.19*** & 0.00 & -1.19*** \\
  \varindent Homosexuality & 2.99 & 2.08 & -0.91*** & 1.96 & -1.03*** \\
  \varindent Prostitution & 2.41 & 0.82 & -1.60*** & 0.77 & -1.64*** \\
  \varindent Abortion & 2.70 & 1.09 & -1.60*** & 0.89 & -1.81*** \\
  \varindent Divorce & 2.86 & 1.24 & -1.62*** & 1.18 & -1.69*** \\
  \varindent Fornication & 2.99 & 1.68 & -1.31*** & 1.51 & -1.48*** \\
  \varindent Suicide & 2.26 & 0.42 & -1.83*** & 0.40 & -1.86*** \\
  \varindent Euthanasia & 3.17 & 1.18 & -1.98*** & 1.05 & -2.11*** \\
  \varindent A man beating his wife & 1.12 & 0.00 & -1.12*** & 0.00 & -1.12*** \\
  \varindent Parents beating children & 2.48 & 0.48 & -1.99*** & 0.49 & -1.99*** \\
  \varindent Violence against other people & 1.35 & 0.00 & -1.35*** & 0.00 & -1.35*** \\
  \varindent Terrorism & 1.12 & 0.00 & -1.12*** & 0.00 & -1.12*** \\
  \varindent Casual sex & 2.60 & 1.16 & -1.44*** & 1.42 & -1.18* \\
  \varindent Political violence & 1.33 & 0.00 & -1.33*** & 0.00 & -1.33*** \\
  \varindent Death penalty & 2.99 & 0.64 & -2.35*** & 0.69 & -2.30*** \\
  \bottomrule
  \end{tabular*}

  \vspace{2pt}
  {\footnotesize \raggedright \setlength{\baselineskip}{9pt} \emph{Note.} Delta is the difference between the silicon sample standard deviation and the WVS standard deviation. Significance is from Fligner-Killeen tests of the equality of variances comparing each silicon sample to the WVS: * $p<0.05$, ** $p<0.01$, *** $p<0.001$.\par}

  \end{table}

\subsubsection{Silicon samples seriously understate opinion variance.}
The divergences between real human populations and silicon samples become much more salient when we look at the dispersion in responses (Table~\ref{tab:sd_comparison}).
Silicon samples \emph{always} underestimate the inherent variance in human opinions.
They have lower dispersion on every single survey item.
Expressed as a share of the WVS standard deviation, silicon samples retain on average only 24\% (slow) and 23\% (fast) of real-world dispersion. 
This is a substantially worse fidelity result than the moderate (${\sim}0.4$ SD) standardized bias found for means in Section~\ref{sec:fidelity-means}.

This is partly attributable to LLMs' tendency to collapse to degenerate homogeneity for survey items assessing attitudes towards serious moral transgressions.
Seven ethical-norm items and one immigration item collapse to an SD of exactly 0.00 in \emph{both} samples.\footnote{The ethical-norm items are: benefits fraud, theft, taking bribes, a man beating his wife, violence against other people, terrorism, and political violence. The immigration item is ``gives the poor a better living.''}
Not \emph{all} items in the ethical norms domain are vulnerable to such collapse.
The models appear to specifically recognize homosexuality as Singapore's one live moral cleavage and represent that contestation with real pluralism, more so than for any other item in the table, with 70\% (slow) and 66\% (fast) of WVS variance preserved among silicon samples.

\subsubsection{Fast sampling and slow sampling are similarly poor at replicating human variance in public opinions.}
Conditional on such poor fidelity, we find little relative difference between fast and slow samples.
The correlation between slow sampling's and fast sampling's SD-deltas across all 27 items is $r=0.984$, and aggregate retained variance is nearly identical between the modes of sampling (23.9\% slow vs.\ 22.9\% fast).
When restricted to the 19 non-degenerate items, we find that fast sampling understates human variance more dramatically than slow sampling on 9 of 12 items on ethical norms.
The obverse is true among immigration items, where slow sampling produces the more homogeneous sample.
However, we note that neither of these between-sampling differences in variance bias are large in absolute terms.

\subsection{How faithful are bivariate associations?}

\begin{landscape}
\begin{figure}
\centering
\caption{Deviation in Cramer's V for immigration attitudes}
\label{fig:immigration_cramerv}
\begin{subfigure}{\linewidth}
  \centering
  \includegraphics[width=\linewidth]{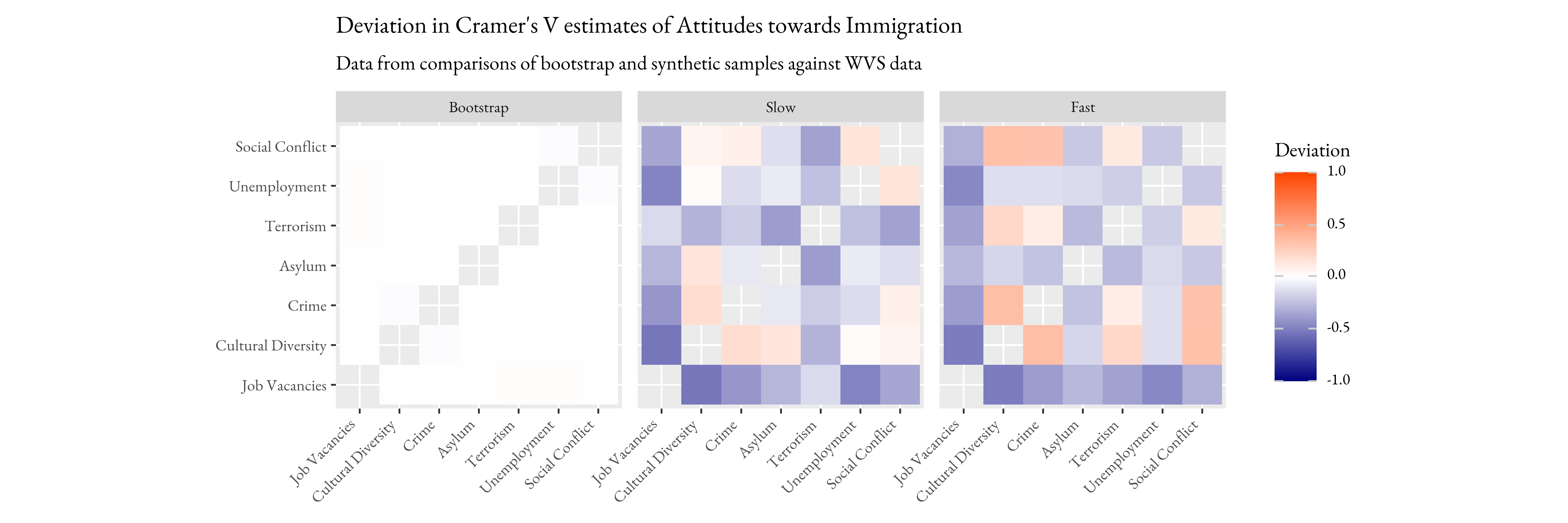}
\end{subfigure}
 
\begin{subfigure}{\linewidth}
  \centering
  \includegraphics[width=\linewidth]{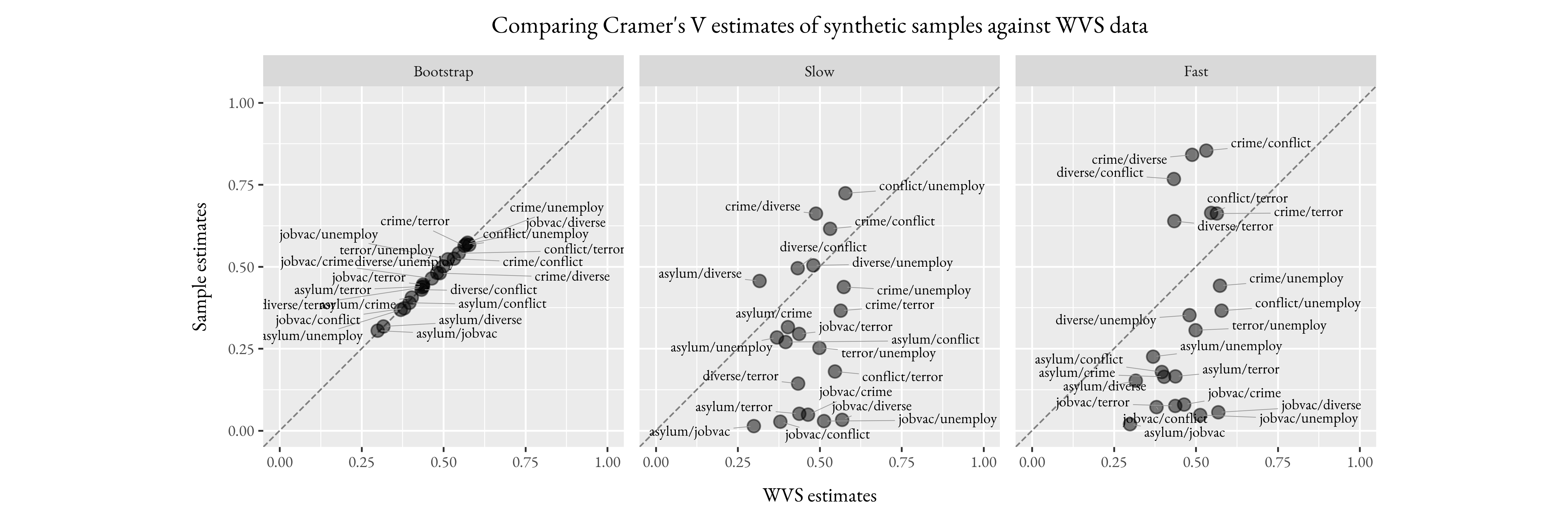}
\end{subfigure}
\end{figure}
\end{landscape}

\begin{landscape}

\begin{figure}
  \centering 
  \caption{Deviation in Cramer's V for moral attitudes}
  \label{fig:ethics_cramerv}
  \begin{subfigure}{\linewidth}
    \centering
    \includegraphics[width=\linewidth]{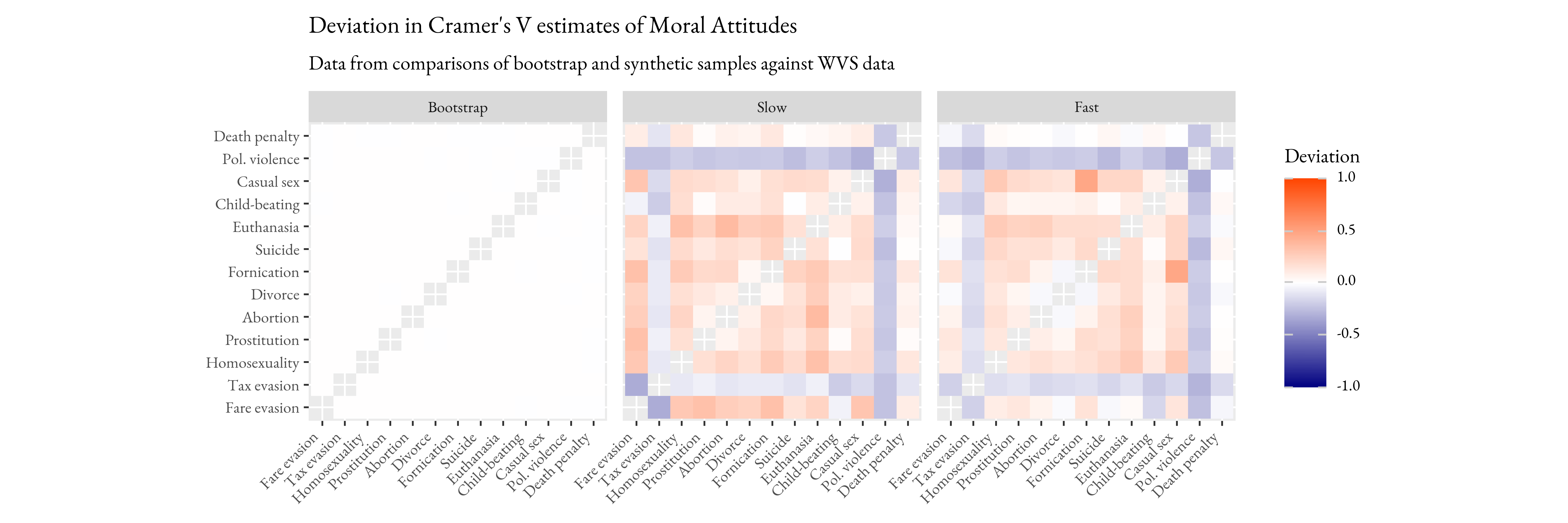}
  \end{subfigure}

  \begin{subfigure}{\linewidth}
    \centering
    \includegraphics[width=\linewidth]{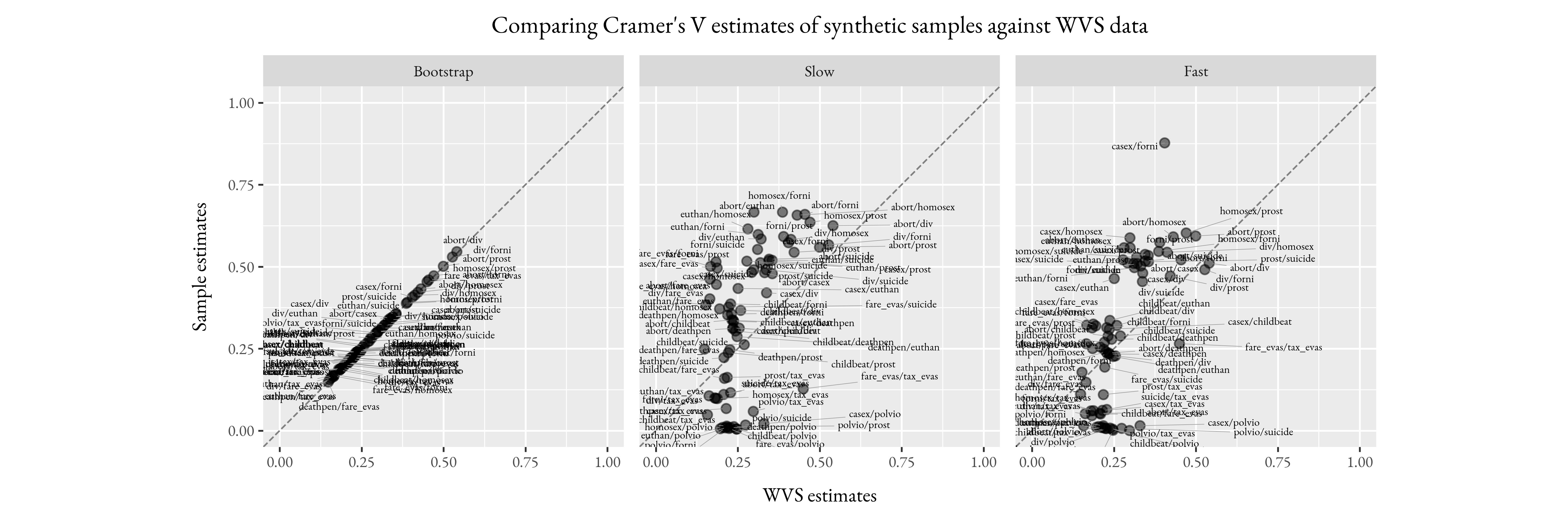}
  \end{subfigure}
\end{figure}
\end{landscape}

To assess whether the silicon samples preserve the bivariate associations among attitudes, we compute the pairwise Cramer's V between each pair of items within each domain. 
Table~\ref{tab:cramersv_immigration} and Table~\ref{tab:cramersv_ethics} (Appendix I) report the resulting matrices for the slow (Panel A) and fast (Panel B) silicon samples, alongside bootstrap samples of the WVS (Panel C) and the WVS benchmark (Panel D); Figures~\ref{fig:immigration_cramerv} and~\ref{fig:ethics_cramerv} visualize the same comparisons. 
Items with no response variation are omitted, since Cramer's V is undefined when a variable is constant. 
Table~\ref{tab:cramersv_summary} summarizes the resulting matrices at the aggregate level; the domain-by-domain and strategy-by-strategy findings below refer back to these figures throughout.

\begin{table}
  \centering
  \caption{Summary Statistics for Pairwise Cramer's V Association Matrices}
  \label{tab:cramersv_summary}
  \footnotesize
  \setbox2=\hbox{%
  \begin{tabular}{lccccccccc}
  \toprule
    & \multicolumn{3}{c}{Correlation with WVS} & \multicolumn{2}{c}{Mean deviation from WVS} & \multicolumn{3}{c}{Slow vs.\ fast} \\
  \cmidrule(lr){2-4} \cmidrule(lr){5-6} \cmidrule(lr){7-9}
    & Slow & Fast & Bootstrap & Slow & Fast & Correlation & Mean $|\Delta|$ & Max $|\Delta|$ \\
  \midrule
  Opinions w.r.t. immigration & 0.28 & 0.42 & 0.998 & $-0.167$ & $-0.123$ & 0.62 & 0.171 & 0.50 \\
  Opinions w.r.t. ethical norms & 0.57 & 0.69 & 1.000 & $+0.060$ & $+0.018$ & 0.93 & 0.066 & 0.30 \\
  \bottomrule
  \end{tabular}}%
  \begin{tabular}{lccccccccc}
  \toprule
    & \multicolumn{3}{c}{Correlation with WVS} & \multicolumn{2}{c}{Mean deviation from WVS} & \multicolumn{3}{c}{Slow vs.\ fast} \\
  \cmidrule(lr){2-4} \cmidrule(lr){5-6} \cmidrule(lr){7-9}
    & Slow & Fast & Bootstrap & Slow & Fast & Correlation & Mean $|\Delta|$ & Max $|\Delta|$ \\
  \midrule
  Opinions w.r.t. immigration & 0.28 & 0.42 & 0.998 & $-0.167$ & $-0.123$ & 0.62 & 0.171 & 0.50 \\
  Opinions w.r.t. ethical norms & 0.57 & 0.69 & 1.000 & $+0.060$ & $+0.018$ & 0.93 & 0.066 & 0.30 \\
  \bottomrule
  \end{tabular}

  \vspace{2pt}
  \parbox{\wd2}{\footnotesize \raggedright \setlength{\baselineskip}{9pt} \emph{Note.} ``Correlation with WVS'' is the Pearson correlation between a sample's full pairwise Cramer's V matrix and the WVS matrix, computed over all unique item pairs with non-degenerate response variation (21 pairs for immigration, 78 for morals); ``Bootstrap'' is a same-population WVS resample, included as a ceiling for the metric. ``Mean deviation from WVS'' is the average, signed difference between a silicon sample's pairwise associations and WVS's (negative = understatement, positive = overstatement). ``Slow vs.\ fast'' reports the same correlation and absolute-deviation statistics computed between the slow and fast silicon samples directly, rather than against WVS.\par}

  \end{table}

\subsubsection{Silicon samples poorly preserve bivariate associations among Singaporean public opinions.}
Associations between attitudes towards immigration are poorly reproduced by silicon samples.
Averaged across all pairs, silicon samples systematically \emph{understate} how tightly immigration attitudes co-vary (mean deviation from WVS of $-0.167$ for slow, $-0.123$ for fast; Table~\ref{tab:cramersv_summary}).
Correlating each silicon sample's (fast or slow) full pairwise association matrix against the WVS matrix gives $r=0.28$ for slow sampling and $r=0.42$ for fast (Table~\ref{tab:cramersv_summary}).
Associations between attitudes towards ethical norms are preserved more faithfully, though still imperfectly.
Averaged across all pairs, silicon samples systematically \emph{overstate} how tightly moral attitudes co-vary, though the average bias is smaller than in the immigration domain and larger for slow than for fast sampling (mean deviation from WVS of $+0.060$ for slow, $+0.018$ for fast; Table~\ref{tab:cramersv_summary}).
Correlating each silicon sample's full pairwise association matrix against the WVS matrix gives $r=0.57$ for slow sampling and $r=0.69$ for fast (Table~\ref{tab:cramersv_summary}).

\subsubsection{Fast sampling is as good as, if not better than, slow sampling.}
Bivariate associations estimated from fast silicon samples are systematically different from those from slow samples.
But, crucially, in both cases they are \emph{not} necessarily less faithful to the WVS estimates than slow silicon samples are.
When it comes to attitudes towards immigration, fast sampling is the more accurate of the two: its higher correlation with WVS ($0.42$ vs.\ $0.28$) and smaller average understatement ($-0.123$ vs.\ $-0.167$) both point the same way (Table~\ref{tab:cramersv_summary}).
The same is true when it comes to attitudes towards ethical norms: fast sampling again correlates more highly with WVS ($0.69$ vs.\ $0.57$) and overstates associations by less on average ($+0.018$ vs.\ $+0.060$), so in both domains and on both metrics fast is the closer of the two to the human benchmark (Table~\ref{tab:cramersv_summary}).
The fast advantage is narrower here, since the two strategies agree far more closely with each other on ethics than on immigration (their association matrices correlate at $r\approx0.93$, versus $r\approx0.63$ on immigration), and it does not hold pair by pair: the largest single slow--fast discrepancy in the domain runs the other way, with fast pushing the casual sex--fornication association to $0.88$ against slow's $0.57$ and a WVS value of $0.40$ (Table~\ref{tab:cramersv_ethics}).

\subsection{How faithful is the contextual space?}

Bivariate associations tell us how tightly any two items move together, but not how the items are arranged relative to one another as a whole.
To assess whether the silicon samples preserve that broader contextual structure, we fit a multiple correspondence analysis (MCA) separately to each source --- slow, fast, bootstrap, and WVS --- within each domain, treating every attitude item as an active categorical variable and retaining the first two dimensions.
Figure~\ref{fig:mca_biplots} plots the resulting configurations for attitudes towards immigration (Panel A) and ethical norms (Panel B).
Unlike the Cramer's V matrices above, items with no response variation are not dropped here; they simply collapse to a single category, which is itself informative and which we take up below.

\begin{landscape}
  \begin{figure}
    \centering
    \caption{Procrustes Alignment of Silicon and WVS MCA Configurations}
    \label{fig:mca_procrustes_overlay}

    \caption*{\textbf{Panel A.} Attitudes towards immigration}
    \includegraphics[width=0.75\linewidth]{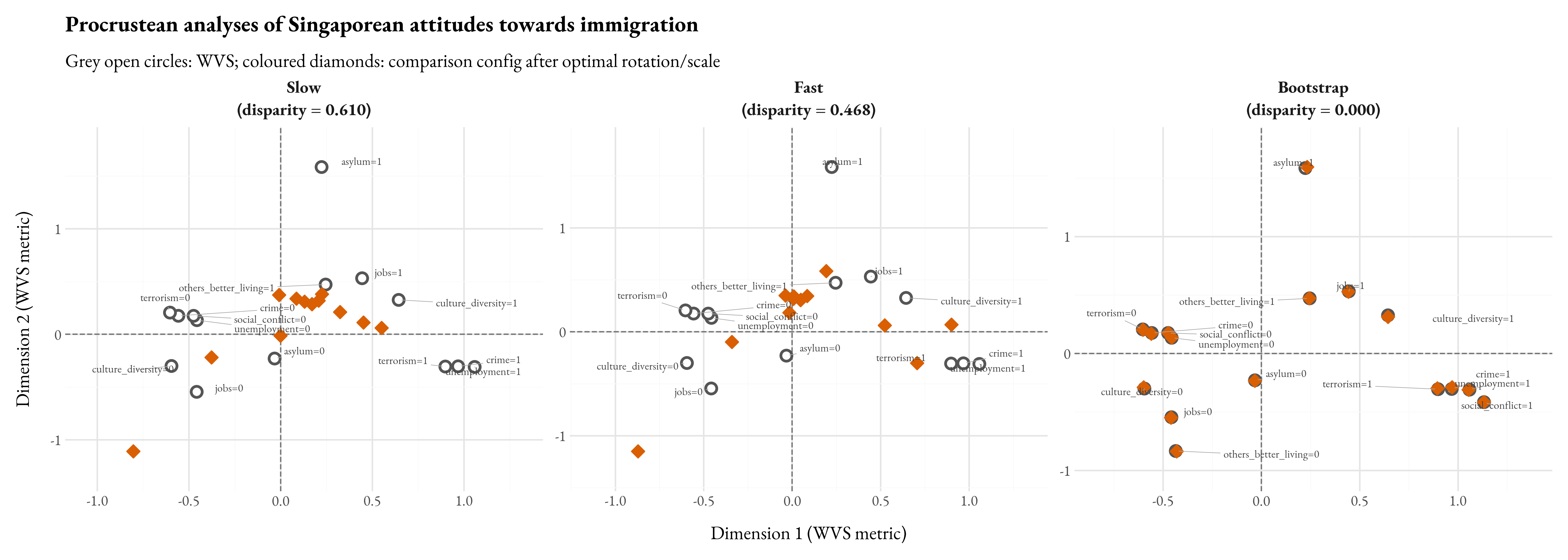}

    \vspace{1em}

    \caption*{\textbf{Panel B.} Attitudes towards ethical norms}
    \includegraphics[width=0.75\linewidth]{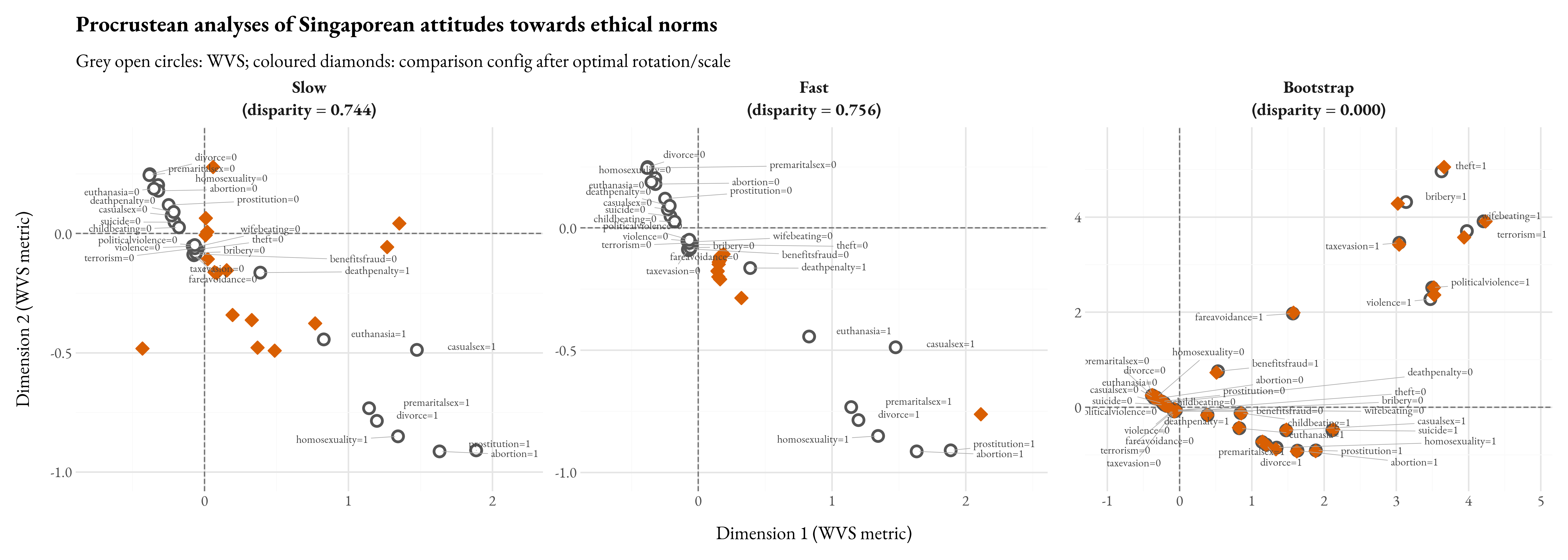}
  \end{figure}
\end{landscape}

Because MCA recovers a configuration only up to translation, rotation, reflection, and scale, the fitted spaces cannot be compared coordinate by coordinate.
We therefore Procrustes-align each silicon configuration onto the WVS solution over the response categories the two share, and measure what disagreement survives that alignment.
Figure~\ref{fig:mca_procrustes_overlay} overlays the aligned configurations and Table~\ref{tab:mca_procrustes} reports the resulting Gower's $M^{2}$ together with Tucker's coefficient of congruence for each dimension, with a same-population WVS bootstrap included throughout as a noise floor.

\begin{landscape}
  \begin{figure}
    \centering
    \caption{MCA Biplots of Attitudes Towards Immigration and Ethical Norms}
    \label{fig:mca_biplots}
    \caption*{\textbf{Panel A.} Attitudes towards immigration}
    \includegraphics[width=0.82\linewidth]{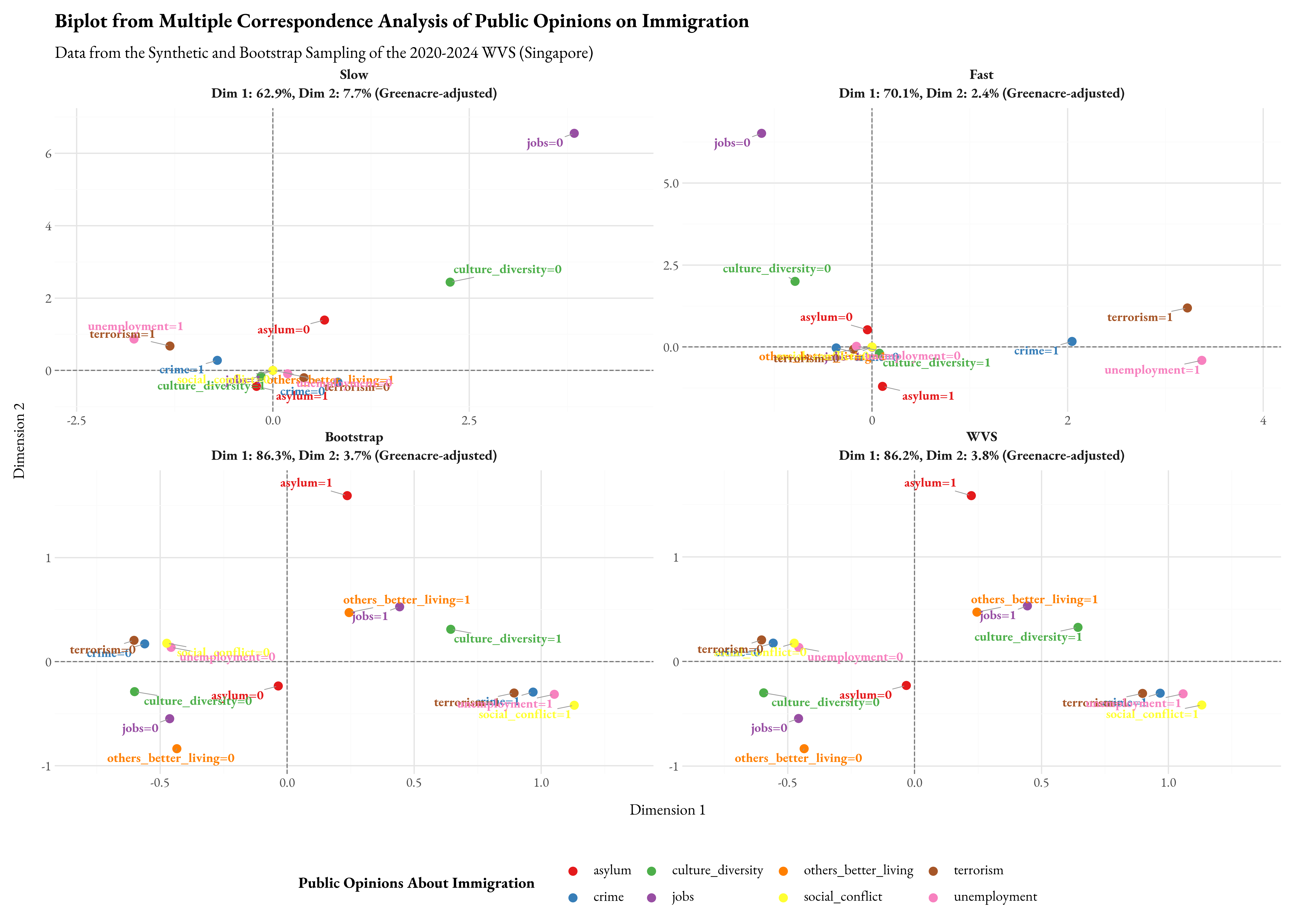}
  \end{figure}
\end{landscape}

\begin{landscape}
  \begin{figure}
    \ContinuedFloat
    \centering
    \caption{MCA Biplots of Attitudes Towards Immigration and Ethical Norms (continued)}
    \caption*{\textbf{Panel B.} Attitudes towards ethical norms}
    \includegraphics[width=0.82\linewidth]{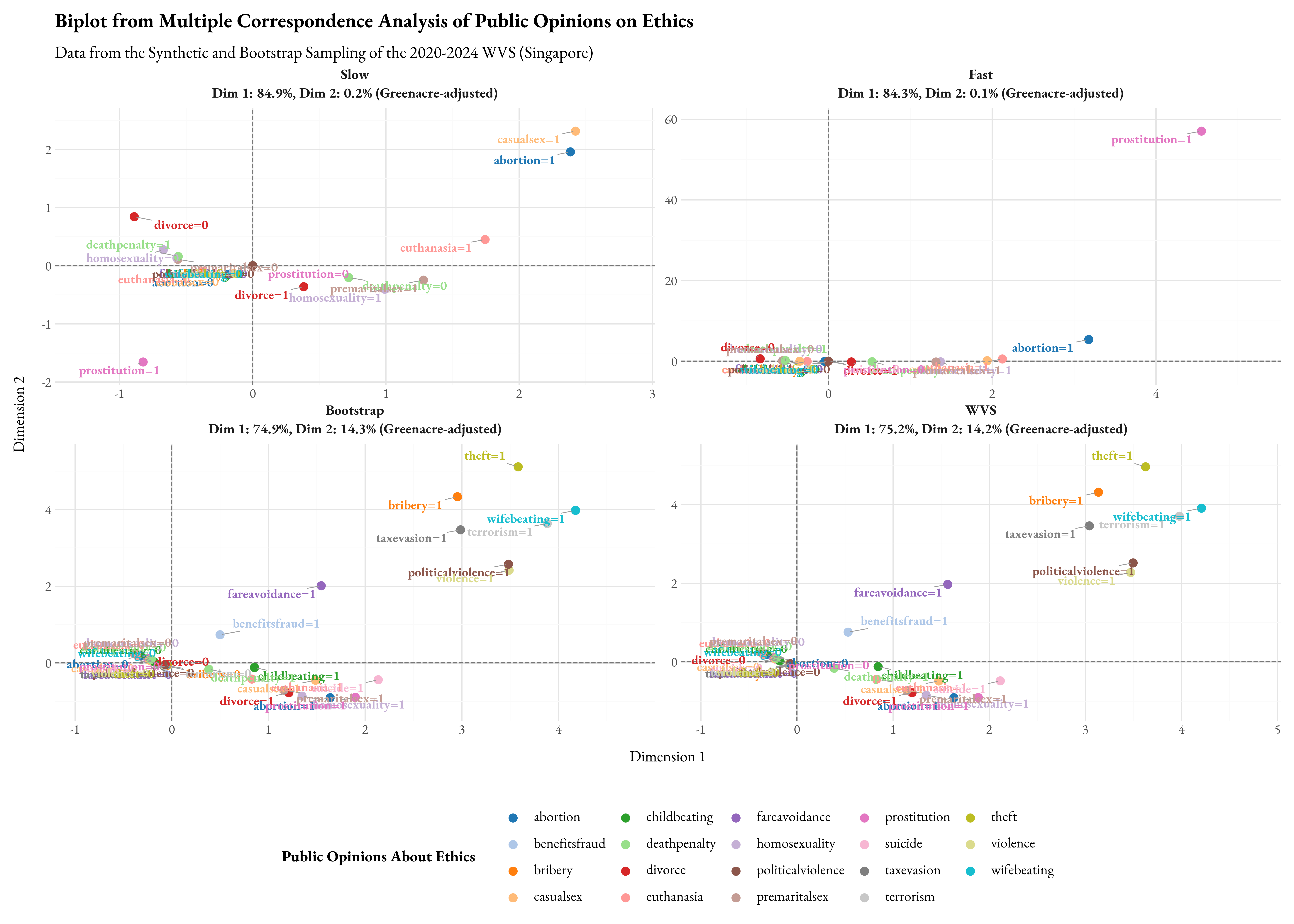}
  \end{figure}
\end{landscape}

\subsubsection{Contextual spaces of immigration and ethical norms are poorly preserved by silicon samples.}
For example, the observed extreme homogeneity among certain ethical norms (Table~\ref{tab:sd_comparison}) shows up directly in the MCA: 11 of the 19 ethics items and 3 of the 8 immigration items collapse to a single category in both silicon samples, leaving no residual variation for these items to relate to anything else.\footnote{We did not observe such trends when looking at bivariate associations because Cramer's V statistics for homogeneous variables are undefined.} 
Table~\ref{tab:mca_procrustes} shows that even the items which do retain some variation are poorly matched to WVS. 
Gower's $M^{2}$ reaches 0.610 (immigration) and 0.744 (ethics) for the slow sample, and 0.468 and 0.756 for the fast one.
Since $M^{2}$ has no conventional cutoffs, the benchmark is the bootstrap, which recovers $M^{2} \leq 0.0004$ under an identical matching procedure (e.g. \citealt{jackson_protest_1995, peres-neto_how_2001}): the silicon residuals are three orders of magnitude above what sampling variation alone produces, and closer to the upper end of the $[0,1]$ range than to the lower.
Tucker's $\varphi$ is correspondingly modest. 
No configuration reaches $0.77$ on either dimension, and the ethics configurations sit below $0.60$ on both, well short of the $0.85$ that \citet{lorenzo-seva_tuckers_2006} take to mark even fair similarity.

\begin{table}
  \centering
  \caption{Procrustes statistics comparing MCA configurations against the WVS}
  \label{tab:mca_procrustes}
  \footnotesize
  \setbox0=\hbox{%
  \begin{tabular}{@{}llrrr@{}}
  \toprule
   & & & \multicolumn{2}{c}{Tucker's $\varphi$} \\
  \cmidrule(lr){4-5}
   & Source & $M^{2}$ & Dim.\ 1 & Dim.\ 2 \\
  \midrule
  Attitudes towards Immigration & Slow & 0.610 & 0.683 & 0.580 \\
   & Fast & 0.468 & 0.769 & 0.688 \\
   & Bootstrap & 0.000 & 1.000 & 1.000 \\
  \midrule
  Ethical Norms & Slow & 0.744 & 0.487 & 0.585 \\
   & Fast & 0.756 & 0.484 & 0.597 \\
   & Bootstrap & 0.000 & 1.000 & 1.000 \\
  \bottomrule
  \end{tabular}}%
  \begin{tabular}{@{}llrrr@{}}
  \toprule
   & & & \multicolumn{2}{c}{Tucker's $\varphi$} \\
  \cmidrule(lr){4-5}
   & Source & $M^{2}$ & Dim.\ 1 & Dim.\ 2 \\
  \midrule
  Attitudes towards Immigration & Slow & 0.610 & 0.683 & 0.580 \\
   & Fast & 0.468 & 0.769 & 0.688 \\
   & Bootstrap & 0.000 & 1.000 & 1.000 \\
  \midrule
  Ethical Norms & Slow & 0.744 & 0.487 & 0.585 \\
   & Fast & 0.756 & 0.484 & 0.597 \\
   & Bootstrap & 0.000 & 1.000 & 1.000 \\
  \bottomrule
  \end{tabular}

  \vspace{2pt}
  \parbox{\wd0}{\footnotesize \raggedright \setlength{\baselineskip}{9pt} \emph{Note.} $M^{2}$ is Gower's Procrustes statistic, a shape-dissimilarity in $[0,1]$ measured after optimal translation, rotation, reflection, and dilation (0~=~identical shapes). Tucker's $\varphi$ is the coefficient of congruence between the WVS and aligned configurations, reported separately for each MCA dimension.}

\end{table}

\subsubsection{Contextual spaces from fast samples resemble those from slow samples.}
Contextual spaces from fast silicon samples are no more different from the WVS than those from slow silicon samples.
Table~\ref{tab:mca_procrustes} shows the two silicon sampling strategies land within a few hundredths of one another on ethics ($M^{2}$ of 0.744 slow vs.\ 0.756 fast; Tucker's $\varphi$ of 0.487/0.585 vs.\ 0.484/0.597 on the two dimensions). 
On immigration the fast sample is if anything the \emph{closer} match to WVS ($M^{2}$ of 0.468 vs.\ 0.610, with $\varphi$ higher on both dimensions, 0.769/0.688 vs.\ 0.683/0.580).

\pagebreak
\hypertarget{conclusion}{%
\section{Conclusion}\label{conclusion}}

Silicon sampling, as of writing, remains an innovative method in early development.
Synthetic surrogates generated by large-language models cannot faithfully reproduce the responses of Singaporean survey respondents on a range of survey questions about immigration and ethical norms.
While silicon samples can provide surprisingly good estimates of mean public opinions, their limitations become obvious when examined more deeply.
Silicon samples do not preserve the higher moments of opinion distribution, and distort the actual covariance and contextual space among public opinions.
These findings are not surprising, and reaffirm past work on earlier generations of large language models (\citealt{bisbee_synthetic_2024,dominguez-olmedo_questioning_2024,santurkar_whose_2023}).
\emph{Conditional} on these limitations, however, we find fast silicon sampling to be relatively superior to slow silicon sampling.
Fast silicon sampling is faster and more efficient with compute than slow sampling; at the same time, it produces synthetic estimates that are as good as, if not more faithful than, those from slow sampling.
While the two modes of silicon sampling broadly share the same deficiencies, we identified no clear instances where fast silicon sampling was notably less faithful than slow silicon sampling.
Indeed, we find some instances where fast sampling, contrary to conventional expectations, is \emph{more faithful} than slow sampling.
When comparing Jaccard distances, synthetic estimates from fast samples are \emph{marginally} closer to WVS estimates than those from slow samples; fast samples are marginally better at reproducing bivariate associations from human data; fast samples are, again, better at preserving the contextual space of attitudes in the immigration domain. 

Our findings carry three important implications for (a) the methodological development of silicon sampling, (b) the present limitations of silicon samples as a method, and (c) the usefulness of geometric analysis to evaluations of relational fidelity of silicon samples. 
One, the conditional superiority of fast silicon sampling over slow silicon sampling.
We find that fast silicon sampling produces samples that are no worse in fidelity than conventional slow sampling methods, and in several diagnostics are marginally better.
The conventional preference for slow, one-call-at-a-time prompting has been justified largely by precedent, by worries about long-context degradation, and by a desire to avoid question-order contamination.
Our results suggest that these worries are, at least for contemporary frontier models and for the domains we study, overstated relative to the costs they impose.
If this pattern is robustly demonstrated by others across models, instruments, and populations, fast silicon sampling ought to become the default convention going forward, given the enormous gains in speed and efficiency.
Such widespread adoption would lower the barrier to constructing large synthetic panels, make iterative prompt and design experiments cheaper to run, and accelerate methodological development further down the line.
The practical upshot is not that silicon sampling suddenly becomes trustworthy, but that when researchers elect to use it, they need not pay the computational premium of slow sampling in order to preserve what little fidelity the method presently affords.

Two, \emph{caveat emptor} remains the message of the day for silicon sampling, especially with regards to public opinions of Singaporeans.
Silicon sampling is very much a method in development, with limited fidelity to real Singaporean samples.
Means can look reassuringly close; higher moments, pairwise associations, and the latent contextual space of attitudes do not.
For Singapore specifically, where training data coverage and cultural representation remain thin relative to Anglo-European survey contexts, the gap between silicon and human samples is not a peripheral nuisance but a first-order threat to inference.
Until fidelity improves on dispersion and relational structure, silicon samples of Singaporean opinion should be treated as exploratory instruments at best -- useful for hypothesis generation, stress-testing survey wording, or bounding model behavior, but not as substitutes for human respondents in substantive claims about what Singaporeans think, believe, or do.

Three, we show how geometric analysis can be used to supplement model evaluation, by providing a clear procedure for measuring divergences in relational fidelity.
Item-level means and even pairwise association matrices leave invisible a form of failure that matters sociologically: a silicon sample can get many margins approximately right while assembling those margins into an opinion space that no human population would recognize.
Multiple correspondence analysis, followed by Procrustes comparison against a human reference configuration, makes that failure measurable rather than merely interpretive.
In our case, the geometric diagnostics expose a sharp divergence that would have been easy to miss from means alone, and they discriminate between sampling modes where bivariate summaries are more ambiguous.
We see this as a portable addition to the evaluation toolkit for silicon sampling that can produce a quick assessment of relational fidelity.

\pagebreak
\bibliographystyle{ACM-Reference-Format}
\bibliography{xy_ref}

\clearpage
\hypertarget{appendix-i}{%
\section{Appendix I: Descriptive Statistics}\label{appendix-i}}

\begin{center}
  \footnotesize
  \captionof{table}{Summary statistics for key demographic covariates}
  \label{tab:wvs_covariate_summary}
\end{center}
\footnotesize
\setlength{\tabcolsep}{4pt}
{\centering
\begin{tabularx}{0.88\linewidth}{Yrrrr}
\toprule
 & Mean & SD & Min & Max \\
\midrule
\varindent $N$ & 2{,}012 & & & \\
\varindent Age & 47.78 & 16.23 & 21 & 91 \\
\varindent Woman & 0.54 & 0.50 & 0 & 1 \\
\varindent Immigrant & 0.20 & 0.40 & 0 & 1 \\
\varindent Citizen & 0.91 & 0.29 & 0 & 1 \\
\varindent Home language is English & 0.43 & 0.49 & 0 & 1 \\
\varindent Currently married & 0.59 & 0.49 & 0 & 1 \\
\varindent Number of children & 1.35 & 1.35 & 0 & 10 \\
\varindent Years of education & 12.25 & 3.90 & 4 & 20 \\
\varindent Parent's years of education & 7.93 & 4.22 & 4 & 20 \\
\varindent Currently employed & 0.64 & 0.48 & 0 & 1 \\
\varindent Subjective class identity & 2.63 & 0.98 & 0 & 5 \\
\varindent Not religious & 0.23 & 0.42 & 0 & 1 \\
\varindent Chinese & 0.77 & 0.42 & 0 & 1 \\
\varindent Malay & 0.12 & 0.32 & 0 & 1 \\
\varindent South Asian & 0.09 & 0.29 & 0 & 1 \\
\bottomrule
\end{tabularx}
\par}

\clearpage

\begin{center}
  \footnotesize
  \captionof{table}{Summary statistics for WVS attitude variables}
  \label{tab:wvs_attitude_summary}
\end{center}
\footnotesize
\setlength{\tabcolsep}{4pt}
{\centering
\begin{tabularx}{0.88\linewidth}{Yrrrr}
\toprule
 & Mean & SD & Min & Max \\
\midrule
\multicolumn{5}{l}{\textbf{Immigration}} \\
\addlinespace[2pt]
\varindent Fills impt. job vacancies & 1.24 & 0.87 & -2 & 2 \\
\varindent Strengths cultural diversity & 1.20 & 0.87 & -2 & 2 \\
\varindent Decreases crime & 1.04 & 0.86 & -2 & 2 \\
\varindent Provides asylum to the persecuted & 0.52 & 0.82 & -2 & 2 \\
\varindent Decreases terrorism risk & 1.10 & 0.88 & -2 & 2 \\
\varindent Gives the poor a better living & 1.48 & 0.79 & -2 & 2 \\
\varindent Decreases unemployment & 0.84 & 0.88 & -2 & 2 \\
\varindent Decreases social conflict & 0.89 & 0.87 & -2 & 2 \\
\addlinespace[4pt]
\midrule
\multicolumn{5}{l}{\textbf{Ethical norms}} \\
\addlinespace[2pt]
\varindent Benefits fraud & 2.51 & 2.20 & -2 & 10 \\
\varindent Fare evasion on public transport & 1.80 & 1.66 & -2 & 10 \\
\varindent Theft & 1.33 & 1.11 & -2 & 10 \\
\varindent Tax evasion & 1.51 & 1.34 & -2 & 10 \\
\varindent Taking bribes & 1.39 & 1.19 & -2 & 10 \\
\varindent Homosexuality & 3.38 & 2.99 & -2 & 10 \\
\varindent Prostitution & 2.73 & 2.41 & -2 & 10 \\
\varindent Abortion & 3.22 & 2.70 & -2 & 10 \\
\varindent Divorce & 4.14 & 2.86 & -2 & 10 \\
\varindent Fornication & 3.96 & 2.99 & -2 & 10 \\
\varindent Suicide & 2.42 & 2.26 & -2 & 10 \\
\varindent Euthanasia & 4.06 & 3.17 & -2 & 10 \\
\varindent A man beating his wife & 1.30 & 1.12 & -2 & 10 \\
\varindent Parents beating children & 3.33 & 2.48 & -2 & 10 \\
\varindent Violence against other people & 1.57 & 1.35 & -2 & 10 \\
\varindent Terrorism & 1.29 & 1.12 & -2 & 10 \\
\varindent Casual sex & 2.81 & 2.60 & -2 & 10 \\
\varindent Political violence & 1.53 & 1.33 & -2 & 10 \\
\varindent Death penalty & 4.70 & 2.99 & -2 & 10 \\
\bottomrule
\end{tabularx}
\par}

\clearpage
\hypertarget{appendix-ii}{%
\section{Appendix II: Cramer's V}\label{appendix-ii}}

\begin{center}
  \footnotesize
  \captionof{table}{Cramer's V between immigration attitude items (all iterations pooled)}
  \label{tab:cramersv_immigration}
\end{center}
\footnotesize

{\centering\textbf{Panel A. Slow silicon sampling}\par}
\vspace{2pt}
      {\centering
      \begin{tabular}[t]{lrrrrrrr}
      \toprule
      & \rotatebox{90}{Job vac.} & \rotatebox{90}{Cul. divers.} & \rotatebox{90}{Crim.} & \rotatebox{90}{Asyl.} & \rotatebox{90}{Terror.} & \rotatebox{90}{Unemploy.} & \rotatebox{90}{Social con.} \\
      \midrule
      Fills impt. job vacancies &  & 0.03 & 0.05 & 0.01 & 0.29 & 0.03 & 0.03 \\
      Strengthens cultural diversity & 0.03 &  & 0.66 & 0.46 & 0.14 & 0.50 & 0.50 \\
      Decreases crime & 0.05 & 0.66 &  & 0.32 & 0.37 & 0.44 & 0.62 \\
      Provides asylum to the persecuted & 0.01 & 0.46 & 0.32 &  & 0.05 & 0.28 & 0.27 \\
      Decreases terrorism risk & 0.29 & 0.14 & 0.37 & 0.05 &  & 0.25 & 0.18 \\
      Decreases unemployment & 0.03 & 0.50 & 0.44 & 0.28 & 0.25 &  & 0.72 \\
      Decreases social conflict & 0.03 & 0.50 & 0.62 & 0.27 & 0.18 & 0.72 &  \\
      \bottomrule
      \end{tabular}
      \par}
      \vspace{18pt}
      {\centering\textbf{Panel B. Fast silicon sampling}\par}
      \vspace{2pt}
      {\centering
      \begin{tabular}[t]{lrrrrrrr}
      \toprule
      & \rotatebox{90}{Job vac.} & \rotatebox{90}{Cul. divers.} & \rotatebox{90}{Crim.} & \rotatebox{90}{Asyl.} & \rotatebox{90}{Terror.} & \rotatebox{90}{Unemploy.} & \rotatebox{90}{Social con.} \\
      \midrule
      Fills impt. job vacancies &  & 0.06 & 0.08 & 0.02 & 0.08 & 0.05 & 0.07 \\
      Strengthens cultural diversity & 0.06 &  & 0.84 & 0.15 & 0.64 & 0.35 & 0.77 \\
      Decreases crime & 0.08 & 0.84 &  & 0.16 & 0.66 & 0.44 & 0.85 \\
      Provides asylum to the persecuted & 0.02 & 0.15 & 0.16 &  & 0.17 & 0.23 & 0.18 \\
      Decreases terrorism risk & 0.08 & 0.64 & 0.66 & 0.17 &  & 0.31 & 0.66 \\
      Decreases unemployment & 0.05 & 0.35 & 0.44 & 0.23 & 0.31 &  & 0.37 \\
      Decreases social conflict & 0.07 & 0.77 & 0.85 & 0.18 & 0.66 & 0.37 &  \\
      \bottomrule
      \end{tabular}
      \par}

\clearpage

{\centering\textbf{Panel C. Bootstrap samples}\par}
      \vspace{2pt}
      {\centering
      \begin{tabular}[t]{lrrrrrrr}
      \toprule
      & \rotatebox{90}{Job vac.} & \rotatebox{90}{Cul. divers.} & \rotatebox{90}{Crim.} & \rotatebox{90}{Asyl.} & \rotatebox{90}{Terror.} & \rotatebox{90}{Unemploy.} & \rotatebox{90}{Social con.} \\
      \midrule
      Fills impt. job vacancies &  & 0.57 & 0.46 & 0.30 & 0.45 & 0.52 & 0.37 \\
      Strengthens cultural diversity & 0.57 &  & 0.48 & 0.32 & 0.44 & 0.48 & 0.43 \\
      Decreases crime & 0.46 & 0.48 &  & 0.41 & 0.56 & 0.57 & 0.52 \\
      Provides asylum to the persecuted & 0.30 & 0.32 & 0.41 &  & 0.44 & 0.37 & 0.39 \\
      Decreases terrorism risk & 0.45 & 0.44 & 0.56 & 0.44 &  & 0.50 & 0.54 \\
      Decreases unemployment & 0.52 & 0.48 & 0.57 & 0.37 & 0.50 &  & 0.57 \\
      Decreases social conflict & 0.37 & 0.43 & 0.52 & 0.39 & 0.54 & 0.57 &  \\
      \bottomrule
      \end{tabular}
      \par}
  
      \vspace{18pt}
      {\centering\textbf{Panel D. World Values Survey (WVS)}\par}
      \vspace{2pt}
      {\centering
      \begin{tabular}[t]{lrrrrrrr}
      \toprule
      & \rotatebox{90}{Job vac.} & \rotatebox{90}{Cul. divers.} & \rotatebox{90}{Crim.} & \rotatebox{90}{Asyl.} & \rotatebox{90}{Terror.} & \rotatebox{90}{Unemploy.} & \rotatebox{90}{Social con.} \\
      \midrule
      Fills impt. job vacancies &  & 0.57 & 0.46 & 0.30 & 0.44 & 0.51 & 0.38 \\
      Strengthens cultural diversity & 0.57 &  & 0.49 & 0.32 & 0.43 & 0.48 & 0.43 \\
      Decreases crime & 0.46 & 0.49 &  & 0.40 & 0.56 & 0.57 & 0.53 \\
      Provides asylum to the persecuted & 0.30 & 0.32 & 0.40 &  & 0.44 & 0.37 & 0.40 \\
      Decreases terrorism risk & 0.44 & 0.43 & 0.56 & 0.44 &  & 0.50 & 0.55 \\
      Decreases unemployment & 0.51 & 0.48 & 0.57 & 0.37 & 0.50 &  & 0.58 \\
      Decreases social conflict & 0.38 & 0.43 & 0.53 & 0.40 & 0.55 & 0.58 &  \\
      \bottomrule
      \end{tabular}
      \par}

\clearpage
\begin{landscape}
\begin{center}
  \footnotesize
  \captionof{table}{Cramer's V between moral attitude items (all iterations pooled)}
  \label{tab:cramersv_ethics}
\end{center}
\footnotesize

{\centering\textbf{Panel A. Slow silicon sampling}\par}
\vspace{2pt}
          {\centering
          \begin{tabular}[t]{lrrrrrrrrrrrrr}
          \toprule
          & \rotatebox{90}{Fare evasion} & \rotatebox{90}{Tax evasion} & \rotatebox{90}{Homosexuality} & \rotatebox{90}{Prostitution} & \rotatebox{90}{Abortion} & \rotatebox{90}{Divorce} & \rotatebox{90}{Fornication} & \rotatebox{90}{Suicide} & \rotatebox{90}{Euthanasia} & \rotatebox{90}{Child-beating} & \rotatebox{90}{Casual sex} & \rotatebox{90}{Pol. violence} & \rotatebox{90}{Death penalty} \\
          \midrule
          Fare evasion &  & 0.13 & 0.46 & 0.51 & 0.45 & 0.40 & 0.50 & 0.38 & 0.38 & 0.16 & 0.50 & 0.00 & 0.25 \\
          Tax evasion & 0.13 &  & 0.10 & 0.16 & 0.11 & 0.10 & 0.10 & 0.12 & 0.11 & 0.01 & 0.07 & 0.06 & 0.05 \\
          Homosexuality & 0.46 & 0.10 &  & 0.64 & 0.66 & 0.58 & 0.67 & 0.51 & 0.62 & 0.37 & 0.49 & 0.01 & 0.35 \\
          Prostitution & 0.51 & 0.16 & 0.64 &  & 0.56 & 0.54 & 0.59 & 0.48 & 0.52 & 0.24 & 0.52 & 0.01 & 0.25 \\
          Abortion & 0.45 & 0.11 & 0.66 & 0.56 &  & 0.62 & 0.66 & 0.52 & 0.67 & 0.34 & 0.48 & 0.01 & 0.31 \\
          Divorce & 0.40 & 0.10 & 0.58 & 0.54 & 0.62 &  & 0.57 & 0.49 & 0.58 & 0.37 & 0.42 & 0.01 & 0.31 \\
          Fornication & 0.50 & 0.10 & 0.67 & 0.59 & 0.66 & 0.57 &  & 0.55 & 0.60 & 0.39 & 0.57 & 0.01 & 0.36 \\
          Suicide & 0.38 & 0.12 & 0.51 & 0.48 & 0.52 & 0.49 & 0.55 &  & 0.52 & 0.26 & 0.48 & 0.02 & 0.22 \\
          Euthanasia & 0.38 & 0.11 & 0.62 & 0.52 & 0.67 & 0.58 & 0.60 & 0.52 &  & 0.33 & 0.43 & 0.01 & 0.29 \\
          Parents beating children & 0.16 & 0.01 & 0.37 & 0.24 & 0.34 & 0.37 & 0.39 & 0.26 & 0.33 &  & 0.31 & 0.01 & 0.30 \\
          Casual sex & 0.50 & 0.07 & 0.49 & 0.52 & 0.48 & 0.42 & 0.57 & 0.48 & 0.43 & 0.31 &  & 0.02 & 0.34 \\
          Political violence & 0.00 & 0.06 & 0.01 & 0.01 & 0.01 & 0.01 & 0.01 & 0.02 & 0.01 & 0.01 & 0.02 &  & 0.01 \\
          Death penalty & 0.25 & 0.05 & 0.35 & 0.25 & 0.31 & 0.31 & 0.36 & 0.22 & 0.29 & 0.30 & 0.34 & 0.01 &  \\
          \bottomrule
          \end{tabular}
          \par}
      
          \newpage
      
          {\centering\textbf{Panel B. Fast silicon sampling}\par}
          \vspace{2pt}
          {\centering
          \begin{tabular}[t]{lrrrrrrrrrrrrr}
          \toprule
          & \rotatebox{90}{Fare evasion} & \rotatebox{90}{Tax evasion} & \rotatebox{90}{Homosexuality} & \rotatebox{90}{Prostitution} & \rotatebox{90}{Abortion} & \rotatebox{90}{Divorce} & \rotatebox{90}{Fornication} & \rotatebox{90}{Suicide} & \rotatebox{90}{Euthanasia} & \rotatebox{90}{Child-beating} & \rotatebox{90}{Casual sex} & \rotatebox{90}{Pol. violence} & \rotatebox{90}{Death penalty} \\
          \midrule
          Fare evasion &  & 0.27 & 0.26 & 0.31 & 0.25 & 0.15 & 0.32 & 0.19 & 0.18 & 0.05 & 0.32 & 0.00 & 0.11 \\
          Tax evasion & 0.27 &  & 0.06 & 0.11 & 0.05 & 0.05 & 0.06 & 0.06 & 0.05 & 0.01 & 0.06 & 0.00 & 0.01 \\
          Homosexuality & 0.26 & 0.06 &  & 0.60 & 0.59 & 0.54 & 0.55 & 0.53 & 0.56 & 0.32 & 0.59 & 0.01 & 0.25 \\
          Prostitution & 0.31 & 0.11 & 0.60 &  & 0.59 & 0.47 & 0.57 & 0.52 & 0.52 & 0.27 & 0.54 & 0.01 & 0.24 \\
          Abortion & 0.25 & 0.05 & 0.59 & 0.59 &  & 0.51 & 0.52 & 0.52 & 0.56 & 0.29 & 0.50 & 0.01 & 0.24 \\
          Divorce & 0.15 & 0.05 & 0.54 & 0.47 & 0.51 &  & 0.49 & 0.45 & 0.51 & 0.32 & 0.48 & 0.01 & 0.23 \\
          Fornication & 0.32 & 0.06 & 0.55 & 0.57 & 0.52 & 0.49 &  & 0.51 & 0.50 & 0.31 & 0.88 & 0.01 & 0.23 \\
          Suicide & 0.19 & 0.06 & 0.53 & 0.52 & 0.52 & 0.45 & 0.51 &  & 0.53 & 0.29 & 0.51 & 0.01 & 0.25 \\
          Euthanasia & 0.18 & 0.05 & 0.56 & 0.52 & 0.56 & 0.51 & 0.50 & 0.53 &  & 0.34 & 0.46 & 0.01 & 0.23 \\
          Parents beating children & 0.05 & 0.01 & 0.32 & 0.27 & 0.29 & 0.32 & 0.31 & 0.29 & 0.34 &  & 0.31 & 0.00 & 0.28 \\
          Casual sex & 0.32 & 0.06 & 0.59 & 0.54 & 0.50 & 0.48 & 0.88 & 0.51 & 0.46 & 0.31 &  & 0.01 & 0.23 \\
          Political violence & 0.00 & 0.00 & 0.01 & 0.01 & 0.01 & 0.01 & 0.01 & 0.01 & 0.01 & 0.00 & 0.01 &  & 0.01 \\
          Death penalty & 0.11 & 0.01 & 0.25 & 0.24 & 0.24 & 0.23 & 0.23 & 0.25 & 0.23 & 0.28 & 0.23 & 0.01 &  \\
          \bottomrule
          \end{tabular}
          \par}
      
          \newpage
      
          {\centering\textbf{Panel C. Bootstrap samples}\par}
          \vspace{2pt}
          {\centering
          \begin{tabular}[t]{lrrrrrrrrrrrrr}
          \toprule
          & \rotatebox{90}{Fare evasion} & \rotatebox{90}{Tax evasion} & \rotatebox{90}{Homosexuality} & \rotatebox{90}{Prostitution} & \rotatebox{90}{Abortion} & \rotatebox{90}{Divorce} & \rotatebox{90}{Fornication} & \rotatebox{90}{Suicide} & \rotatebox{90}{Euthanasia} & \rotatebox{90}{Child-beating} & \rotatebox{90}{Casual sex} & \rotatebox{90}{Pol. violence} & \rotatebox{90}{Death penalty} \\
          \midrule
          Fare evasion &  & 0.45 & 0.17 & 0.19 & 0.19 & 0.16 & 0.17 & 0.23 & 0.16 & 0.21 & 0.19 & 0.25 & 0.15 \\
          Tax evasion & 0.45 &  & 0.19 & 0.22 & 0.21 & 0.18 & 0.19 & 0.23 & 0.17 & 0.22 & 0.22 & 0.30 & 0.16 \\
          Homosexuality & 0.17 & 0.19 &  & 0.47 & 0.43 & 0.41 & 0.39 & 0.32 & 0.28 & 0.19 & 0.30 & 0.21 & 0.22 \\
          Prostitution & 0.19 & 0.22 & 0.47 &  & 0.50 & 0.42 & 0.39 & 0.36 & 0.28 & 0.22 & 0.35 & 0.24 & 0.22 \\
          Abortion & 0.19 & 0.21 & 0.43 & 0.50 &  & 0.55 & 0.46 & 0.35 & 0.30 & 0.24 & 0.34 & 0.21 & 0.24 \\
          Divorce & 0.16 & 0.18 & 0.41 & 0.42 & 0.55 &  & 0.53 & 0.34 & 0.32 & 0.26 & 0.34 & 0.22 & 0.26 \\
          Fornication & 0.17 & 0.19 & 0.39 & 0.39 & 0.46 & 0.53 &  & 0.31 & 0.31 & 0.23 & 0.41 & 0.22 & 0.23 \\
          Suicide & 0.23 & 0.23 & 0.32 & 0.36 & 0.35 & 0.34 & 0.31 &  & 0.35 & 0.27 & 0.30 & 0.27 & 0.21 \\
          Euthanasia & 0.16 & 0.17 & 0.28 & 0.28 & 0.30 & 0.32 & 0.31 & 0.35 &  & 0.24 & 0.25 & 0.20 & 0.24 \\
          Parents beating children & 0.21 & 0.22 & 0.19 & 0.22 & 0.24 & 0.26 & 0.23 & 0.27 & 0.24 &  & 0.24 & 0.24 & 0.25 \\
          Casual sex & 0.19 & 0.22 & 0.30 & 0.35 & 0.34 & 0.34 & 0.41 & 0.30 & 0.25 & 0.24 &  & 0.33 & 0.24 \\
          Political violence & 0.25 & 0.30 & 0.21 & 0.24 & 0.21 & 0.22 & 0.22 & 0.27 & 0.20 & 0.24 & 0.33 &  & 0.23 \\
          Death penalty & 0.15 & 0.16 & 0.22 & 0.22 & 0.24 & 0.26 & 0.23 & 0.21 & 0.24 & 0.25 & 0.24 & 0.23 &  \\
          \bottomrule
          \end{tabular}
          \par}
      
          \newpage
      
          {\centering\textbf{Panel D. World Values Survey (WVS)}\par}
          \vspace{2pt}
          {\centering
          \begin{tabular}[t]{lrrrrrrrrrrrrr}
          \toprule
          & \rotatebox{90}{Fare evasion} & \rotatebox{90}{Tax evasion} & \rotatebox{90}{Homosexuality} & \rotatebox{90}{Prostitution} & \rotatebox{90}{Abortion} & \rotatebox{90}{Divorce} & \rotatebox{90}{Fornication} & \rotatebox{90}{Suicide} & \rotatebox{90}{Euthanasia} & \rotatebox{90}{Child-beating} & \rotatebox{90}{Casual sex} & \rotatebox{90}{Pol. violence} & \rotatebox{90}{Death penalty} \\
          \midrule
          Fare evasion &  & 0.45 & 0.17 & 0.18 & 0.18 & 0.16 & 0.17 & 0.22 & 0.15 & 0.21 & 0.19 & 0.25 & 0.15 \\
          Tax evasion & 0.45 &  & 0.19 & 0.22 & 0.21 & 0.18 & 0.18 & 0.23 & 0.16 & 0.22 & 0.21 & 0.30 & 0.16 \\
          Homosexuality & 0.17 & 0.19 &  & 0.47 & 0.43 & 0.41 & 0.39 & 0.32 & 0.28 & 0.19 & 0.30 & 0.20 & 0.22 \\
          Prostitution & 0.18 & 0.22 & 0.47 &  & 0.50 & 0.42 & 0.39 & 0.36 & 0.28 & 0.22 & 0.34 & 0.24 & 0.23 \\
          Abortion & 0.18 & 0.21 & 0.43 & 0.50 &  & 0.54 & 0.45 & 0.35 & 0.30 & 0.23 & 0.33 & 0.21 & 0.23 \\
          Divorce & 0.16 & 0.18 & 0.41 & 0.42 & 0.54 &  & 0.53 & 0.34 & 0.32 & 0.26 & 0.34 & 0.22 & 0.25 \\
          Fornication & 0.17 & 0.18 & 0.39 & 0.39 & 0.45 & 0.53 &  & 0.31 & 0.31 & 0.23 & 0.40 & 0.22 & 0.23 \\
          Suicide & 0.22 & 0.23 & 0.32 & 0.36 & 0.35 & 0.34 & 0.31 &  & 0.35 & 0.27 & 0.29 & 0.27 & 0.21 \\
          Euthanasia & 0.15 & 0.16 & 0.28 & 0.28 & 0.30 & 0.32 & 0.31 & 0.35 &  & 0.24 & 0.25 & 0.20 & 0.25 \\
          Parents beating children & 0.21 & 0.22 & 0.19 & 0.22 & 0.23 & 0.26 & 0.23 & 0.27 & 0.24 &  & 0.24 & 0.25 & 0.24 \\
          Casual sex & 0.19 & 0.21 & 0.30 & 0.34 & 0.33 & 0.34 & 0.40 & 0.29 & 0.25 & 0.24 &  & 0.33 & 0.24 \\
          Political violence & 0.25 & 0.30 & 0.20 & 0.24 & 0.21 & 0.22 & 0.22 & 0.27 & 0.20 & 0.25 & 0.33 &  & 0.23 \\
          Death penalty & 0.15 & 0.16 & 0.22 & 0.23 & 0.23 & 0.25 & 0.23 & 0.21 & 0.25 & 0.24 & 0.24 & 0.23 &  \\
          \bottomrule
          \end{tabular}
          \par}
        \end{landscape}

\clearpage
\hypertarget{appendix-iii}{%
\section{Appendix III: MCA Contributions}\label{appendix-iii}}

\begin{center}
  \footnotesize
  \captionof{table}{MCA category contributions for immigration attitude items.}
  \label{tab:mca_ctr_immigration}
\end{center}
\footnotesize
{\centering\textbf{Panel A. Slow silicon sampling}\par}
\vspace{2pt}
  {\centering
  \begin{tabular}[t]{@{}llrrrrrr@{}}
  \toprule
  & & \multicolumn{3}{c}{Dimension 1} & \multicolumn{3}{c}{Dimension 2} \\
  \cmidrule(lr){3-5} \cmidrule(lr){6-8}
  Item & Cat. & Coord. & Ctr (\%) & $\cos^2$ & Coord. & Ctr (\%) & $\cos^2$ \\
  \midrule
  Fills impt. job vacancies & 0 & 3.84 & 0.10 & 0.00 & 6.55 & 0.43 & 0.01 \\
   & 1 & 0.00 & 0.00 & 0.00 & 0.00 & 0.00 & 0.01 \\
  Strengthens cultural diversity & 0 & 2.26 & 16.56 & 0.34 & 2.44 & 28.02 & 0.39 \\
   & 1 & -0.15 & 1.10 & 0.34 & -0.16 & 1.86 & 0.39 \\
  Decreases crime & 0 & 0.83 & 16.51 & 0.59 & -0.32 & 3.68 & 0.09 \\
   & 1 & -0.71 & 14.11 & 0.59 & 0.28 & 3.14 & 0.09 \\
  Provides asylum to the persecuted & 0 & 0.66 & 5.44 & 0.14 & 1.39 & 35.29 & 0.61 \\
   & 1 & -0.21 & 1.73 & 0.14 & -0.44 & 11.20 & 0.61 \\
  Decreases terrorism risk & 0 & 0.40 & 6.28 & 0.52 & -0.20 & 2.40 & 0.14 \\
   & 1 & -1.31 & 20.81 & 0.52 & 0.67 & 7.95 & 0.14 \\
  Gives the poor a better living & 1 & 0.00 & 0.00 & --- & 0.00 & 0.00 & --- \\
  Decreases unemployment & 0 & 0.19 & 1.67 & 0.33 & -0.09 & 0.58 & 0.08 \\
   & 1 & -1.77 & 15.69 & 0.33 & 0.87 & 5.46 & 0.08 \\
  Decreases social conflict & 0 & 0.00 & 0.00 & --- & 0.00 & 0.00 & --- \\
  \bottomrule
  \end{tabular}
  \par}
\vspace{18pt}
{\centering\textbf{Panel B. Fast silicon sampling}\par}
\vspace{2pt}
  {\centering
  \begin{tabular}[t]{@{}llrrrrrr@{}}
  \toprule
  & & \multicolumn{3}{c}{Dimension 1} & \multicolumn{3}{c}{Dimension 2} \\
  \cmidrule(lr){3-5} \cmidrule(lr){6-8}
  Item & Cat. & Coord. & Ctr (\%) & $\cos^2$ & Coord. & Ctr (\%) & $\cos^2$ \\
  \midrule
  Fills impt. job vacancies & 0 & -1.13 & 0.11 & 0.00 & 6.51 & 5.84 & 0.07 \\
   & 1 & 0.00 & 0.00 & 0.00 & -0.01 & 0.01 & 0.07 \\
  Strengthens cultural diversity & 0 & -0.79 & 2.77 & 0.06 & 2.00 & 29.78 & 0.38 \\
   & 1 & 0.08 & 0.27 & 0.06 & -0.19 & 2.87 & 0.38 \\
  Decreases crime & 0 & -0.37 & 5.81 & 0.75 & -0.03 & 0.06 & 0.01 \\
   & 1 & 2.05 & 32.42 & 0.75 & 0.17 & 0.36 & 0.01 \\
  Provides asylum to the persecuted & 0 & -0.05 & 0.08 & 0.01 & 0.52 & 16.20 & 0.63 \\
   & 1 & 0.11 & 0.18 & 0.01 & -1.20 & 37.21 & 0.63 \\
  Decreases terrorism risk & 0 & -0.19 & 1.70 & 0.61 & -0.07 & 0.38 & 0.08 \\
   & 1 & 3.23 & 29.22 & 0.61 & 1.19 & 6.60 & 0.08 \\
  Gives the poor a better living & 1 & 0.00 & 0.00 & --- & 0.00 & 0.00 & --- \\
  Decreases unemployment & 0 & -0.16 & 1.24 & 0.54 & 0.02 & 0.03 & 0.01 \\
   & 1 & 3.37 & 26.21 & 0.54 & -0.41 & 0.64 & 0.01 \\
  Decreases social conflict & 0 & 0.00 & 0.00 & --- & 0.00 & 0.00 & --- \\
  \bottomrule
  \end{tabular}
  \par}
\clearpage
{\centering\textbf{Panel C. Bootstrap samples}\par}
\vspace{2pt}
  {\centering
  \begin{tabular}[t]{@{}llrrrrrr@{}}
  \toprule
  & & \multicolumn{3}{c}{Dimension 1} & \multicolumn{3}{c}{Dimension 2} \\
  \cmidrule(lr){3-5} \cmidrule(lr){6-8}
  Item & Cat. & Coord. & Ctr (\%) & $\cos^2$ & Coord. & Ctr (\%) & $\cos^2$ \\
  \midrule
  Fills impt. job vacancies & 0 & -0.46 & 3.73 & 0.20 & -0.55 & 10.73 & 0.29 \\
   & 1 & 0.44 & 3.57 & 0.20 & 0.53 & 10.28 & 0.29 \\
  Strengthens cultural diversity & 0 & -0.60 & 6.66 & 0.39 & -0.29 & 3.15 & 0.09 \\
   & 1 & 0.64 & 7.14 & 0.39 & 0.31 & 3.37 & 0.09 \\
  Decreases crime & 0 & -0.56 & 7.10 & 0.54 & 0.17 & 1.33 & 0.05 \\
   & 1 & 0.97 & 12.26 & 0.54 & -0.29 & 2.30 & 0.05 \\
  Provides asylum to the persecuted & 0 & -0.03 & 0.04 & 0.01 & -0.23 & 3.49 & 0.37 \\
   & 1 & 0.24 & 0.26 & 0.01 & 1.59 & 23.71 & 0.37 \\
  Decreases terrorism risk & 0 & -0.60 & 7.74 & 0.54 & 0.20 & 1.80 & 0.06 \\
   & 1 & 0.89 & 11.47 & 0.54 & -0.30 & 2.68 & 0.06 \\
  Gives the poor a better living & 0 & -0.43 & 2.42 & 0.11 & -0.84 & 18.35 & 0.39 \\
   & 1 & 0.24 & 1.36 & 0.11 & 0.47 & 10.32 & 0.39 \\
  Decreases unemployment & 0 & -0.46 & 5.19 & 0.48 & 0.14 & 0.95 & 0.04 \\
   & 1 & 1.05 & 11.95 & 0.48 & -0.31 & 2.18 & 0.04 \\
  Decreases social conflict & 0 & -0.47 & 5.64 & 0.54 & 0.18 & 1.59 & 0.07 \\
   & 1 & 1.13 & 13.46 & 0.54 & -0.42 & 3.79 & 0.07 \\
  \bottomrule
  \end{tabular}
  \par}
\vspace{18pt}
{\centering\textbf{Panel D. World Values Survey (WVS)}\par}
\vspace{2pt}
  {\centering
  \begin{tabular}[t]{@{}llrrrrrr@{}}
  \toprule
  & & \multicolumn{3}{c}{Dimension 1} & \multicolumn{3}{c}{Dimension 2} \\
  \cmidrule(lr){3-5} \cmidrule(lr){6-8}
  Item & Cat. & Coord. & Ctr (\%) & $\cos^2$ & Coord. & Ctr (\%) & $\cos^2$ \\
  \midrule
  Fills impt. job vacancies & 0 & -0.46 & 3.68 & 0.20 & -0.55 & 10.68 & 0.29 \\
   & 1 & 0.44 & 3.58 & 0.20 & 0.53 & 10.39 & 0.29 \\
  Strengthens cultural diversity & 0 & -0.59 & 6.56 & 0.38 & -0.30 & 3.42 & 0.10 \\
   & 1 & 0.64 & 7.11 & 0.38 & 0.33 & 3.70 & 0.10 \\
  Decreases crime & 0 & -0.56 & 7.03 & 0.54 & 0.17 & 1.41 & 0.05 \\
   & 1 & 0.97 & 12.22 & 0.54 & -0.30 & 2.45 & 0.05 \\
  Provides asylum to the persecuted & 0 & -0.03 & 0.03 & 0.01 & -0.23 & 3.36 & 0.36 \\
   & 1 & 0.22 & 0.23 & 0.01 & 1.58 & 23.14 & 0.36 \\
  Decreases terrorism risk & 0 & -0.60 & 7.77 & 0.54 & 0.21 & 1.83 & 0.06 \\
   & 1 & 0.90 & 11.57 & 0.54 & -0.31 & 2.73 & 0.06 \\
  Gives the poor a better living & 0 & -0.44 & 2.44 & 0.11 & -0.83 & 18.27 & 0.39 \\
   & 1 & 0.25 & 1.38 & 0.11 & 0.47 & 10.31 & 0.39 \\
  Decreases unemployment & 0 & -0.46 & 5.19 & 0.48 & 0.13 & 0.91 & 0.04 \\
   & 1 & 1.06 & 12.04 & 0.48 & -0.31 & 2.10 & 0.04 \\
  Decreases social conflict & 0 & -0.47 & 5.66 & 0.54 & 0.17 & 1.56 & 0.07 \\
   & 1 & 1.13 & 13.51 & 0.54 & -0.42 & 3.73 & 0.07 \\
  \bottomrule
  \end{tabular}
  \par}

\clearpage

  \begin{center}
    \scriptsize
    \captionof{table}{MCA category contributions for moral attitude items.}
    \label{tab:mca_ctr_ethics}
  \end{center}
  \scriptsize
  {\centering\textbf{Panel A. Slow silicon sampling}\par}
  \vspace{2pt}
    {\centering
    \begin{tabular}[t]{@{}llrrrrrr@{}}
    \toprule
    & & \multicolumn{3}{c}{Dimension 1} & \multicolumn{3}{c}{Dimension 2} \\
    \cmidrule(lr){3-5} \cmidrule(lr){6-8}
    Item & Cat. & Coord. & Ctr (\%) & $\cos^2$ & Coord. & Ctr (\%) & $\cos^2$ \\
    \midrule
    Benefits fraud & 0 & 0.00 & 0.00 & 0.21 & 0.00 & 0.00 & 0.14 \\
    Fare evasion & 0 & 0.00 & 0.00 & 0.21 & 0.00 & 0.00 & 0.14 \\
    Theft & 0 & 0.00 & 0.00 & 0.21 & 0.00 & 0.00 & 0.14 \\
    Tax evasion & 0 & 0.00 & 0.00 & 0.21 & 0.00 & 0.00 & 0.14 \\
    Bribery & 0 & 0.00 & 0.00 & 0.21 & 0.00 & 0.00 & 0.14 \\
    Homosexuality & 0 & -0.67 & 7.55 & 0.67 & 0.27 & 4.00 & 0.11 \\
     & 1 & 0.99 & 11.16 & 0.67 & -0.41 & 5.91 & 0.11 \\
    Prostitution & 0 & 0.00 & 0.00 & 0.00 & 0.00 & 0.00 & 0.00 \\
     & 1 & -0.82 & 0.00 & 0.00 & -1.66 & 0.00 & 0.00 \\
    Abortion & 0 & -0.18 & 0.85 & 0.43 & -0.15 & 1.82 & 0.29 \\
     & 1 & 2.39 & 11.27 & 0.43 & 1.96 & 24.05 & 0.29 \\
    Divorce & 0 & -0.89 & 6.70 & 0.34 & 0.84 & 19.02 & 0.31 \\
     & 1 & 0.38 & 2.89 & 0.34 & -0.36 & 8.20 & 0.31 \\
    Premarital sex & 0 & -0.56 & 6.19 & 0.72 & 0.11 & 0.74 & 0.03 \\
     & 1 & 1.28 & 14.10 & 0.72 & -0.25 & 1.68 & 0.03 \\
    Suicide & 0 & 0.00 & 0.00 & 0.21 & 0.00 & 0.00 & 0.14 \\
    Euthanasia & 0 & -0.37 & 3.22 & 0.65 & -0.10 & 0.68 & 0.04 \\
     & 1 & 1.75 & 15.08 & 0.65 & 0.45 & 3.17 & 0.04 \\
    Wife-beating & 0 & 0.00 & 0.00 & 0.21 & 0.00 & 0.00 & 0.14 \\
    Child-beating & 0 & 0.00 & 0.00 & 0.21 & 0.00 & 0.00 & 0.14 \\
    Violence & 0 & 0.00 & 0.00 & 0.21 & 0.00 & 0.00 & 0.14 \\
    Terrorism & 0 & 0.00 & 0.00 & 0.21 & 0.00 & 0.00 & 0.14 \\
    Casual sex & 0 & -0.14 & 0.53 & 0.34 & -0.13 & 1.54 & 0.31 \\
     & 1 & 2.42 & 9.11 & 0.34 & 2.31 & 26.32 & 0.31 \\
    Political violence & 0 & 0.00 & 0.00 & 0.21 & 0.00 & 0.00 & 0.14 \\
    Death penalty & 0 & 0.72 & 6.38 & 0.40 & -0.20 & 1.62 & 0.03 \\
     & 1 & -0.56 & 4.95 & 0.40 & 0.16 & 1.26 & 0.03 \\
    \bottomrule
    \end{tabular}
    \par}
  \vspace{18pt}
  {\centering\textbf{Panel B. Fast silicon sampling}\par}
  \vspace{2pt}
    {\centering
    \begin{tabular}[t]{@{}llrrrrrr@{}}
    \toprule
    & & \multicolumn{3}{c}{Dimension 1} & \multicolumn{3}{c}{Dimension 2} \\
    \cmidrule(lr){3-5} \cmidrule(lr){6-8}
    Item & Cat. & Coord. & Ctr (\%) & $\cos^2$ & Coord. & Ctr (\%) & $\cos^2$ \\
    \midrule
    Benefits fraud & 0 & 0.00 & 0.00 & 0.10 & 0.00 & 0.00 & --- \\
    Fare evasion & 0 & 0.00 & 0.00 & 0.10 & 0.00 & 0.00 & --- \\
    Theft & 0 & 0.00 & 0.00 & 0.10 & 0.00 & 0.00 & --- \\
    Tax evasion & 0 & 0.00 & 0.00 & 0.10 & 0.00 & 0.00 & --- \\
    Bribery & 0 & 0.00 & 0.00 & 0.10 & 0.00 & 0.00 & --- \\
    Homosexuality & 0 & -0.54 & 6.14 & 0.73 & 0.07 & 0.33 & 0.01 \\
     & 1 & 1.37 & 15.71 & 0.73 & -0.18 & 0.83 & 0.01 \\
    Prostitution & 0 & 0.00 & 0.00 & 0.00 & -0.01 & 0.01 & 0.49 \\
     & 1 & 4.56 & 0.09 & 0.00 & 57.04 & 44.42 & 0.49 \\
    Abortion & 0 & -0.04 & 0.06 & 0.14 & -0.08 & 0.51 & 0.40 \\
     & 1 & 3.18 & 4.18 & 0.14 & 5.35 & 36.48 & 0.40 \\
    Divorce & 0 & -0.83 & 5.22 & 0.23 & 0.57 & 7.64 & 0.11 \\
     & 1 & 0.28 & 1.76 & 0.23 & -0.19 & 2.58 & 0.11 \\
    Premarital sex & 0 & -0.56 & 6.54 & 0.74 & 0.09 & 0.52 & 0.02 \\
     & 1 & 1.31 & 15.36 & 0.74 & -0.21 & 1.22 & 0.02 \\
    Suicide & 0 & 0.00 & 0.00 & 0.10 & 0.00 & 0.00 & --- \\
    Euthanasia & 0 & -0.26 & 1.79 & 0.55 & -0.07 & 0.35 & 0.04 \\
     & 1 & 2.12 & 14.63 & 0.55 & 0.54 & 2.88 & 0.04 \\
    Wife-beating & 0 & 0.00 & 0.00 & 0.10 & 0.00 & 0.00 & --- \\
    Child-beating & 0 & 0.00 & 0.00 & 0.10 & 0.00 & 0.00 & --- \\
    Violence & 0 & 0.00 & 0.00 & 0.10 & 0.00 & 0.00 & --- \\
    Terrorism & 0 & 0.00 & 0.00 & 0.10 & 0.00 & 0.00 & --- \\
    Casual sex & 0 & -0.35 & 3.07 & 0.68 & -0.02 & 0.03 & 0.00 \\
     & 1 & 1.94 & 17.08 & 0.68 & 0.11 & 0.16 & 0.00 \\
    Political violence & 0 & 0.00 & 0.00 & 0.10 & 0.00 & 0.00 & --- \\
    Death penalty & 0 & 0.53 & 4.21 & 0.28 & -0.15 & 1.03 & 0.02 \\
     & 1 & -0.53 & 4.15 & 0.28 & 0.15 & 1.01 & 0.02 \\
    \bottomrule
    \end{tabular}
    \par}
\clearpage
  {\centering\textbf{Panel C. Bootstrap samples}\par}
  \vspace{2pt}
    {\centering
    \begin{tabular}[t]{@{}llrrrrrr@{}}
    \toprule
    & & \multicolumn{3}{c}{Dimension 1} & \multicolumn{3}{c}{Dimension 2} \\
    \cmidrule(lr){3-5} \cmidrule(lr){6-8}
    Item & Cat. & Coord. & Ctr (\%) & $\cos^2$ & Coord. & Ctr (\%) & $\cos^2$ \\
    \midrule
    Benefits fraud & 0 & -0.06 & 0.05 & 0.03 & -0.08 & 0.22 & 0.06 \\
     & 1 & 0.50 & 0.48 & 0.03 & 0.73 & 1.89 & 0.06 \\
    Fare evasion & 0 & -0.07 & 0.09 & 0.11 & -0.09 & 0.29 & 0.19 \\
     & 1 & 1.55 & 1.98 & 0.11 & 2.01 & 6.18 & 0.19 \\
    Theft & 0 & -0.05 & 0.04 & 0.17 & -0.07 & 0.15 & 0.34 \\
     & 1 & 3.58 & 3.07 & 0.17 & 5.11 & 11.52 & 0.34 \\
    Tax evasion & 0 & -0.08 & 0.11 & 0.23 & -0.09 & 0.26 & 0.31 \\
     & 1 & 2.99 & 4.14 & 0.23 & 3.47 & 10.31 & 0.31 \\
    Bribery & 0 & -0.06 & 0.06 & 0.16 & -0.08 & 0.22 & 0.35 \\
     & 1 & 2.96 & 3.00 & 0.16 & 4.33 & 11.90 & 0.35 \\
    Homosexuality & 0 & -0.32 & 1.56 & 0.44 & 0.21 & 1.19 & 0.18 \\
     & 1 & 1.35 & 6.51 & 0.44 & -0.87 & 4.98 & 0.18 \\
    Prostitution & 0 & -0.25 & 1.02 & 0.47 & 0.12 & 0.42 & 0.11 \\
     & 1 & 1.90 & 7.78 & 0.47 & -0.90 & 3.22 & 0.11 \\
    Abortion & 0 & -0.32 & 1.61 & 0.53 & 0.18 & 0.91 & 0.16 \\
     & 1 & 1.64 & 8.19 & 0.53 & -0.91 & 4.65 & 0.16 \\
    Divorce & 0 & -0.38 & 2.06 & 0.46 & 0.25 & 1.57 & 0.19 \\
     & 1 & 1.21 & 6.54 & 0.46 & -0.78 & 4.98 & 0.19 \\
    Premarital sex & 0 & -0.39 & 2.09 & 0.45 & 0.24 & 1.45 & 0.17 \\
     & 1 & 1.16 & 6.26 & 0.45 & -0.71 & 4.35 & 0.17 \\
    Suicide & 0 & -0.21 & 0.74 & 0.45 & 0.04 & 0.06 & 0.02 \\
     & 1 & 2.14 & 7.56 & 0.45 & -0.44 & 0.59 & 0.02 \\
    Euthanasia & 0 & -0.35 & 1.63 & 0.29 & 0.18 & 0.78 & 0.08 \\
     & 1 & 0.83 & 3.82 & 0.29 & -0.42 & 1.83 & 0.08 \\
    Wife-beating & 0 & -0.06 & 0.07 & 0.25 & -0.06 & 0.11 & 0.23 \\
     & 1 & 4.18 & 4.61 & 0.25 & 3.97 & 7.71 & 0.23 \\
    Child-beating & 0 & -0.18 & 0.50 & 0.15 & 0.03 & 0.02 & 0.00 \\
     & 1 & 0.86 & 2.37 & 0.15 & -0.12 & 0.09 & 0.00 \\
    Violence & 0 & -0.08 & 0.11 & 0.28 & -0.05 & 0.10 & 0.13 \\
     & 1 & 3.49 & 5.01 & 0.28 & 2.41 & 4.43 & 0.13 \\
    Terrorism & 0 & -0.07 & 0.08 & 0.26 & -0.06 & 0.13 & 0.23 \\
     & 1 & 3.89 & 4.71 & 0.26 & 3.63 & 7.61 & 0.23 \\
    Casual sex & 0 & -0.23 & 0.86 & 0.34 & 0.07 & 0.15 & 0.03 \\
     & 1 & 1.49 & 5.53 & 0.34 & -0.45 & 0.95 & 0.03 \\
    Political violence & 0 & -0.07 & 0.08 & 0.23 & -0.05 & 0.08 & 0.12 \\
     & 1 & 3.48 & 4.16 & 0.23 & 2.57 & 4.19 & 0.12 \\
    Death penalty & 0 & -0.21 & 0.54 & 0.08 & 0.09 & 0.18 & 0.01 \\
     & 1 & 0.38 & 0.98 & 0.08 & -0.16 & 0.33 & 0.01 \\
    \bottomrule
    \end{tabular}
    \par}
  \vspace{18pt}
  {\centering\textbf{Panel D. World Values Survey (WVS)}\par}
  \vspace{2pt}
    {\centering
    \begin{tabular}[t]{@{}llrrrrrr@{}}
    \toprule
    & & \multicolumn{3}{c}{Dimension 1} & \multicolumn{3}{c}{Dimension 2} \\
    \cmidrule(lr){3-5} \cmidrule(lr){6-8}
    Item & Cat. & Coord. & Ctr (\%) & $\cos^2$ & Coord. & Ctr (\%) & $\cos^2$ \\
    \midrule
    Benefits fraud & 0 & -0.06 & 0.06 & 0.03 & -0.09 & 0.23 & 0.06 \\
     & 1 & 0.53 & 0.54 & 0.03 & 0.75 & 2.01 & 0.06 \\
    Fare evasion & 0 & -0.07 & 0.09 & 0.11 & -0.09 & 0.27 & 0.18 \\
     & 1 & 1.57 & 2.02 & 0.11 & 1.97 & 5.89 & 0.18 \\
    Theft & 0 & -0.05 & 0.04 & 0.18 & -0.07 & 0.15 & 0.33 \\
     & 1 & 3.63 & 3.27 & 0.18 & 4.96 & 11.34 & 0.33 \\
    Tax evasion & 0 & -0.08 & 0.11 & 0.24 & -0.09 & 0.26 & 0.30 \\
     & 1 & 3.04 & 4.26 & 0.24 & 3.46 & 10.21 & 0.30 \\
    Bribery & 0 & -0.06 & 0.06 & 0.18 & -0.08 & 0.22 & 0.35 \\
     & 1 & 3.14 & 3.35 & 0.18 & 4.31 & 11.76 & 0.35 \\
    Homosexuality & 0 & -0.32 & 1.54 & 0.43 & 0.20 & 1.15 & 0.17 \\
     & 1 & 1.34 & 6.46 & 0.43 & -0.85 & 4.80 & 0.17 \\
    Prostitution & 0 & -0.25 & 1.01 & 0.47 & 0.12 & 0.44 & 0.11 \\
     & 1 & 1.89 & 7.67 & 0.47 & -0.91 & 3.31 & 0.11 \\
    Abortion & 0 & -0.32 & 1.59 & 0.52 & 0.18 & 0.92 & 0.16 \\
     & 1 & 1.63 & 8.10 & 0.52 & -0.91 & 4.71 & 0.16 \\
    Divorce & 0 & -0.38 & 2.02 & 0.45 & 0.25 & 1.62 & 0.20 \\
     & 1 & 1.20 & 6.39 & 0.45 & -0.79 & 5.11 & 0.20 \\
    Premarital sex & 0 & -0.38 & 2.03 & 0.44 & 0.24 & 1.55 & 0.18 \\
     & 1 & 1.14 & 6.07 & 0.44 & -0.73 & 4.63 & 0.18 \\
    Suicide & 0 & -0.21 & 0.73 & 0.44 & 0.05 & 0.07 & 0.02 \\
     & 1 & 2.12 & 7.42 & 0.44 & -0.49 & 0.72 & 0.02 \\
    Euthanasia & 0 & -0.35 & 1.60 & 0.29 & 0.19 & 0.85 & 0.08 \\
     & 1 & 0.83 & 3.78 & 0.29 & -0.44 & 2.02 & 0.08 \\
    Wife-beating & 0 & -0.06 & 0.07 & 0.26 & -0.06 & 0.11 & 0.22 \\
     & 1 & 4.21 & 4.73 & 0.26 & 3.91 & 7.57 & 0.22 \\
    Child-beating & 0 & -0.18 & 0.48 & 0.15 & 0.03 & 0.02 & 0.00 \\
     & 1 & 0.84 & 2.29 & 0.15 & -0.12 & 0.09 & 0.00 \\
    Violence & 0 & -0.08 & 0.12 & 0.28 & -0.05 & 0.09 & 0.12 \\
     & 1 & 3.47 & 5.11 & 0.28 & 2.27 & 4.06 & 0.12 \\
    Terrorism & 0 & -0.07 & 0.08 & 0.26 & -0.06 & 0.13 & 0.23 \\
     & 1 & 3.98 & 4.81 & 0.26 & 3.70 & 7.73 & 0.23 \\
    Casual sex & 0 & -0.23 & 0.83 & 0.34 & 0.08 & 0.17 & 0.04 \\
     & 1 & 1.48 & 5.39 & 0.34 & -0.49 & 1.10 & 0.04 \\
    Political violence & 0 & -0.07 & 0.08 & 0.24 & -0.05 & 0.08 & 0.12 \\
     & 1 & 3.50 & 4.28 & 0.24 & 2.51 & 4.11 & 0.12 \\
    Death penalty & 0 & -0.21 & 0.54 & 0.08 & 0.09 & 0.18 & 0.01 \\
     & 1 & 0.39 & 0.99 & 0.08 & -0.16 & 0.33 & 0.01 \\
    \bottomrule
    \end{tabular}
    \par}

\end{document}